\pdfoutput=1
\documentclass{article}
\PassOptionsToPackage{numbers, compress}{natbib}
\usepackage[utf8]{inputenc}
\usepackage[accsupp]{axessibility}  
\usepackage[main=USenglish]{babel}
\usepackage[T1]{fontenc}
\usepackage[final]{neurips_2024}
\usepackage{newtxtext}
\usepackage{textcomp}
\usepackage{amsmath}
\usepackage{amssymb}
\usepackage{adjustbox}
\usepackage{array}
\usepackage{booktabs}
\usepackage{arydshln} 
\usepackage{breakcites}
\usepackage{color}
\usepackage{colortbl}
\usepackage[autostyle]{csquotes}
\usepackage{etoolbox}
\usepackage{graphbox}
\usepackage{graphicx}
\usepackage{makecell}
\usepackage{algorithm}
\usepackage[italicComments=false,noEnd=true]{algpseudocodex}
\usepackage{mathtools}
\usepackage{microtype}
\usepackage{calc}
\usepackage{mleftright}
\usepackage{multirow}
\usepackage{nicefrac}
\usepackage{pgfplots}
\pgfplotsset{compat=1.18}
\usepackage{rotating}
\usepackage{stmaryrd}
\usepackage{setspace}
\usepackage[mode=match]{siunitx}
\usepackage{subcaption}
\usepackage{tabularx}
\usepackage{xfrac}
\definecolor{cvprblue}{rgb}{0.21,0.49,0.74}
\usepackage[pagebackref,verbose=true]{hyperref}
\usepackage{varioref}
\usepackage[capitalize]{cleveref}
\usepackage{marginnote}
\usepackage{todonotes}
\usepackage{titlesec}
\crefname{equation}{}{}
\crefname{section}{Sec.}{Secs.}
\Crefname{section}{Section}{Sections}
\crefname{table}{Tab.}{Tabs.}
\Crefname{table}{Table}{Tables}

\newcommand*{\B}{\bfseries}

\DeclareSIUnit\pixel{px}

\DeclareSIUnit\fps{fps}

\defineshorthand{"~}{\babelhyphen{nobreak}}
\useshorthands{"}

\makeatletter
\newcommand*\titledcaption[2]{\caption{{\textbf{#1\@.}} #2}}
\makeatother



\makeatletter

\renewcommand{\@addpunct}[1]{%
  \ifhmode
    \ifnum\spacefactor>\@m \else#1\fi
  \fi
}

\let\paragraphnoautodot\paragraph
\RenewDocumentCommand{\paragraph}{som}{%
	\IfBooleanTF{#1}
	{\paragraphnoautodot*{#3}}
	{\IfNoValueTF{#2}
		{\paragraphnoautodot{\maybe@addperiod{#3}}}
		{\paragraphnoautodot[#2]{\maybe@addperiod{#3}}}%
	}%
}
\newcommand{\maybe@addperiod}[1]{%
	#1\@addpunct{.}%
}
\makeatother

\makeatletter
\newcommand{\paragraphsentence}[1]{
  \scr@startsection {paragraph}
  {\csname paragraphnumdepth\endcsname }
  {\csname scr@paragraph@sectionindent\endcsname }
  {\csname scr@paragraph@beforeskip\endcsname }
  {0pt}  
  {\ifdim \glueexpr \csname scr@paragraph@afterskip\endcsname >\z@ \expandafter \ifnum \scr@v@is@gt {2.96}\relax \setlength {\parfillskip }{\z@ plus 1fil}\fi \fi \raggedsection \normalfont \sectfont \nobreak \usekomafont {paragraph}}{#1\ }} 
\makeatother

\newcommand{\T}{^T}

\DeclarePairedDelimiter\parenstmp{\lparen}{\rparen}
\DeclarePairedDelimiter\abstmp{\lvert}{\rvert}
\DeclarePairedDelimiter\normtmp{\lVert}{\rVert}
\DeclarePairedDelimiter\floortmp{\lfloor}{\rfloor}
\DeclarePairedDelimiter\ceiltmp{\lceil}{\rceil}
\DeclarePairedDelimiter\bracestmp{\lbrace}{\rbrace}
\DeclarePairedDelimiter\bracketstmp{\lbrack}{\rbrack}
\DeclarePairedDelimiter\anglestmp{\langle}{\rangle}
\NewDocumentCommand{\applyDelimiter}{mmm}{%
  \IfValueTF{#1}
    {#2[#1]{#3}} 
    {#2*{#3}}    
}
\NewDocumentCommand{\parens}{om}{\applyDelimiter{#1}{\parenstmp}{#2}}
\NewDocumentCommand{\abs}{om}{\applyDelimiter{#1}{\abstmp}{#2}}
\NewDocumentCommand{\norm}{om}{\applyDelimiter{#1}{\normtmp}{#2}}
\NewDocumentCommand{\floor}{om}{\applyDelimiter{#1}{\floortmp}{#2}}
\NewDocumentCommand{\ceil}{om}{\applyDelimiter{#1}{\ceiltmp}{#2}}
\NewDocumentCommand{\braces}{om}{\applyDelimiter{#1}{\bracestmp}{#2}}
\NewDocumentCommand{\brackets}{om}{\applyDelimiter{#1}{\bracketstmp}{#2}}
\NewDocumentCommand{\angles}{om}{\applyDelimiter{#1}{\anglestmp}{#2}}

\newcommand{\operatorcall}[2]{\operatorname{#1}\parens{#2}}

\newcommand{\R}{\mathbb{R}}

\newcommand{\N}{\mathbb{N}}
\newcommand{\loss}[1]{\mathcal{L}_\mathrm{#1}}

\mathchardef\mhyphen="2D

\newcommand{\z}{\phantom{0}}
\newcommand{\pr}{^\prime}
\newcommand{\p}{\mathbf{p}}
\newcommand{\phat}{\mathbf{\hat{p}}}
\input{config/abbreviations}

\titlespacing*{\paragraph}{0pt}{1ex plus 1ex minus 0.5ex}{1em}

\definecolor{tabblue}{HTML}{1f77b4}
\definecolor{taborange}{HTML}{ff7f0e}
\definecolor{tabgreen}{HTML}{2ca02c}
\definecolor{tabred}{HTML}{d62728}
\definecolor{tabpurple}{HTML}{9467bd}
\definecolor{tabbrown}{HTML}{8c564b}
\definecolor{tabpink}{HTML}{e377c2}
\definecolor{tabgray}{HTML}{7f7f7f}
\definecolor{tabolive}{HTML}{bcbd22}
\definecolor{tabcyan}{HTML}{17becf}

\hypersetup{
  colorlinks=true,
  breaklinks=true,
  linkcolor=tabred,
  citecolor=tabblue,
  filecolor=tabpink,
  urlcolor=tabcyan,
  plainpages=false,
  pdfencoding=unicode,
  pdfpagemode=UseOutlines,
}

\title{Neural Localizer Fields for Continuous \\ 3D Human Pose and Shape Estimation}
\author{%
  Istv\'an S\'ar\'andi\textmd{,\negmedspace\textsuperscript{1,2}} \ \ \ 
  Gerard Pons-Moll\textmd{\textsuperscript{1,2,3}}\vspace{0.5ex}\\ 
  \textsuperscript{1}University of Tübingen, Germany, \ 
  \textsuperscript{2}Tübingen AI Center, Germany,\\
  \textsuperscript{3}Max Planck Institute for Informatics, Saarland Informatics Campus, Germany\\
  \texttt{\url{https://istvansarandi.com/nlf}}
}

\begin{document}
\graphicspath{{../figures/}}

\maketitle

\begin{abstract}
With the explosive growth of available training data, single-image 3D human modeling is ahead of a transition to a data-centric paradigm.
A key to successfully exploiting data scale is to design flexible models that can be supervised from various heterogeneous data sources produced by different researchers or vendors.
To this end, we propose a simple yet powerful paradigm for seamlessly unifying different human pose and shape-related tasks and datasets.
Our formulation is centered on the ability -- both at training and test time -- to query any arbitrary point of the human volume, and obtain its estimated location in 3D.
We achieve this by learning a continuous neural field of body point localizer functions, each of which is a differently parameterized 3D heatmap-based convolutional point localizer (detector).
For generating parametric output, we propose an efficient post-processing step for fitting SMPL-family body models to nonparametric joint and vertex predictions.
With this approach, we can naturally exploit differently annotated data sources including mesh, 2D/3D skeleton and dense pose, without having to convert between them, and thereby train large-scale 3D human mesh and skeleton estimation models that considerably outperform the state-of-the-art on several public benchmarks including 3DPW, EMDB, EHF, SSP-3D and AGORA.
\end{abstract}

\section{Introduction}
\label{sec:introduction}
Along with the wider field of computer vision, the human pose and shape estimation community has recently started a transition towards leveraging large-scale training data to produce robust models~\cite{Sarandi23WACV,goel2023humans,cai2023smplerx,Pang22NeurIPSDB,lin2023motionx}.
When assembling large training sets from different sources, it is inevitable that they will use different types of annotations, including different parametric mesh models, 3D skeletal joints and markers from different motion capture vendors and manual 2D keypoint annotations.
Re-annotation to a single format for training is a costly and error-prone process, and therefore difficult to scale.
Similarly, current models are trained to output a specific format, such as pre-defined joints, keypoints, or parametric body meshes, which is limiting for downstream applications which require other formats.
We therefore argue that a good model design should be able to ingest and output highly heterogeneous forms of annotations in a unified manner. 

Despite the different kinds of labels, these tasks all share a common underlying factor: the correspondence problem of mapping from a canonical human body space to the observation space of the camera.
Hence, if a model could localize arbitrary human points (both surface and body-internal ones, \eg joints), its predictions could be supervised from any human-centric data source with point labels and could be customized to new landmark sets even after deployment.
In this work, we propose a method that is able to achieve exactly this.

To localize individual points, we leverage the machinery of volumetric heatmap estimation with soft-argmax decoding, which has worked well in skeletal keypoint estimation methods~\cite{Pavlakos17CVPR,Sun18ECCV,Sarandi21TBIOM}.
However, generalizing this to arbitrary points is nontrivial.
In prior work, each body joint heatmap is produced through a separate set of convolution weights, tying the model to a particular set of joints or keypoints.
Extending this paradigm to localize an infinite number of points would require infinite memory to store the weights, and learning would be highly inefficient as weights would need to be trained independently for each point. 

Our key idea and contribution is to learn a continuous field of point localizers, which we call a \emph{Neural Localizer Field}. 
The principle is to modulate a deep network's prediction layer on-the-fly, based on what point we need to predict: at training time querying points for which we have annotations, and at test time sampling the points that an application requires.
Here, we are inspired by the recent success of neural field representations, such as neural radiance~\cite{mildenhall2020nerf}, distance~\cite{park2019deepsdf,chibane2020ndf} and occupancy fields~\cite{Occupancy_Networks}, but instead of predicting a physical property (occupancy, color) we predict a function.

\begin{figure}[t]
\centering
\includegraphics[width=\linewidth]{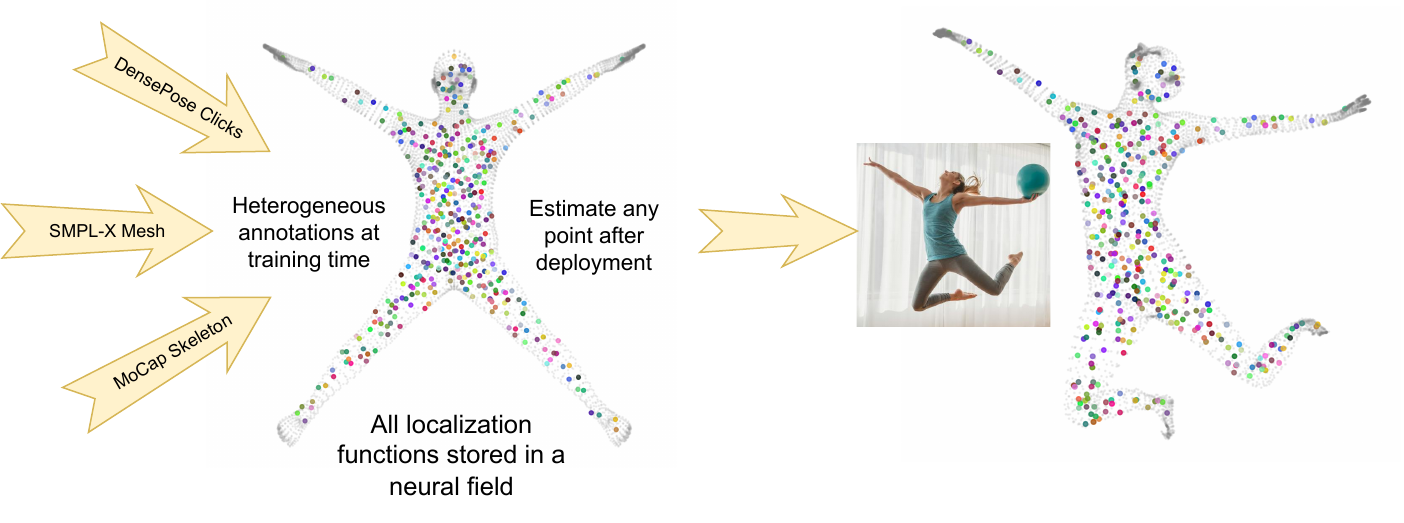}\\
\caption{\emph{Can one model learn to localize any point of the human body in 3D from a single RGB image?} We propose to build a generalist human pose and shape estimator that can readily learn from any annotated points at training time and can estimate any user-chosen points at test time.}
\label{fig:teaser}
\end{figure}

Specifically, this \emph{localizer field}'s input domain is the full 3D volume of the canonical human body and its output range is the parameter space of the modulated convolutional output layer (making the localizer field a hypernetwork~\cite{ha2017hypernetworks,karras2019style,cao2022authentic}).
By optimizing the neural field's parameters, we obtain a smooth field of localizers (functions) over the human volume domain, improving robustness, and knowledge sharing.
In the paper, we investigate various design choices related to this architecture, including positional encodings.
The architecture results in consistent and accurate skeletal joint predictions along with smooth nonparametric mesh output, whose format is chosen by the user at test time, see \cref{fig:teaser}.
Since a parametric representation (for example based on SMPL~\cite{Loper15TOG}) is often preferred for further downstream processing, 
as the second contribution of this paper, we derive a simple and efficient algorithm for fitting SMPL-family body models to our non-parametric predictions. 
This involves alternating between estimating global body part orientations through the Kabsch algorithm~\cite{Kabsch1976ASF}, and solving for the shape coefficients in a regularized linear least squares formulation.
This converges after as few as ca.\@ 2--4 rounds, and is well-suited for GPU acceleration.

In summary, our main contribution is the \emph{Neural Localizer Field} method for the 3D localization of an \emph{infinite continuum of points within the human body volume}, based on a single RGB image, allowing seamless mixed-dataset training using various skeleton or mesh annotation formats without the need for tedious and error-prone re-annotation. 
Through this formulation, we are able to train a generalist human pose and shape estimator that outperforms prior work on a wide range of important benchmarks.
We will make our code and trained models publicly available for research. 

\section{Related Work}
\label{sec:related-work}
Methods for 3D pose/shape estimation use a wide variety of formats: different skeletons, meshes, body models or volumetric representations.
For a comprehensive review, see~\cite{Tian22Arxiv}.
Here we focus on methods that closely relate to ours.

\paragraph{Large-scale multi-dataset human-centric training}
A few recent works~\cite{Sarandi23WACV,goel2023humans,cai2023smplerx} have started to explore scaling up 3D human body reconstruction methods beyond the previously common practice of training on one or a few datasets.
Models trained at scale from many sources tend to outperform specialized models that were trained for a specific dataset.
However, current methods have to unify their datasets to a common format~\cite{cai2023smplerx}, which is often an ill-posed problem.
This can introduce errors if done through pseudo-annotation, requiring significant human effort and manual quality checks.
Another option is to train a model to predict all possible outputs and optimize consistency losses between them~\cite{Sarandi23WACV}, but this is not scalable to a large, possibly infinite number of annotation types.
By stark contrast, our model can ingest data from many different formats, and can be asked to output any point set at test time.

\paragraph{Nonparametric mesh estimation and sparse keypoints}
Some works~\cite{Lassner17CVPR,ma20233d} predict further keypoints on the surface beyond the skeleton, noting that this carries richer shape information.
Ma~\etal~\cite{ma20233d} predict a sparse set of virtual markers and approximate the rest through convex interpolation.
Nonparametric mesh estimation has been a successful line of recent research~\cite{cho2022FastMETRO,yoshiyasu2023deformable,Lin21CVPR,Lin21ICCV,Corona22ECCV,choi2020pose2mesh,kolotouros2019cmr,moon2020i2l,luan2021pc}, mainly due to better preservation of alignment in contrast to parametric estimation, which often suffers from mean-shape bias~\cite{STRAPS2020BMVC}.
In contrast to our work, the prominent Transformer-based approaches of this category~\cite{Lin21CVPR,cho2022FastMETRO,Lin21ICCV} feed a static set of keypoints into their architecture throughout training, which then exchange information through self-attention layers, and require a coarse to fine strategy and complex design for efficient processing.
Our architecture is more lightweight, and can directly produce predictions for all possible points, without having to interpolate them from some other set of points.

\paragraph{Continuous and dense representations}
While the dominant paradigm to supervise pose estimators has been sparse keypoints and skeletons, a few works have explored dense, continuously varying keypoints for supervision and prediction.
DenseReg~\cite{alp2017densereg} predicts a dense 2D deformation grid to self-supervise a face regressor.
In DensePose~\cite{Guler18CVPR,neverova2020continuous}, 2D keypoint estimation is generalized to a continuous representation, where arbitrary surface points can be regression targets, instead of having a fixed set of sparse keypoints throughout the dataset.
DensePose3D~\cite{shapovalov2021densepose} shows how 2D dense maps can be used to lift estimates to 3D.
Similar in spirit, DenseBody~\cite{yao2019densebody,zeng20203d} predicts 3D mesh points on a 2D UV-map.
Chandran~\etal~\cite{chandran2023continuous} estimate 2D facial landmarks in a continuous manner.
Common to all these works is that either the prediction or some intermediate representations are given in 2D, \eg UV-maps, dense semantic maps or 2D output coordinates.
By contrast, we are able to perform a direct 3D-to-3D mapping from canonical to observation space, to localize an infinite continuum of points in the human volume, to harness heterogeneous annotations.

\paragraph{Distinctions from DensePose}
DensePose~\cite{Guler18CVPR,neverova2020continuous} may seem similar enough to our motivation that we should expand upon why it does not solve the same problem we are tackling.
In a sense, our problem formulation is the \emph{reverse} of DensePose's.
DensePose starts out in image space and asks, \emph{per pixel}, what canonical surface point is visible at the pixel (facing the camera).
This inherently restricts the formulation to the frontal surface of the body.
Occluded points, body-internal joints and the opposite-side surface of the human are ignored in DensePose.
Instead, we start in canonical space and ask the more natural question of where each body point (identified in canonical space) can be found in 3D observation space.
This is more in line with the goals in practical applications.
However, because of this reverse relation between the two formulations, the annotations in DensePose datasets~\cite{Neverova18ECCV,Guler18CVPR} can still be directly used to supervise our model as well (with the roles reversed), further boosting the data sources available to us.

\paragraph{3D keypoint estimation through heatmaps}
A well-established paradigm to predict 3D keypoints is to use 3D heatmaps~\cite{Pavlakos17CVPR,Sun18ECCV,Sarandi21TBIOM,ma20233d,Mehta20TOG,Mehta17TOG}.
The advantage is that convolutional architectures are well suited for localizing joints in the image, and inputs and outputs are well aligned.
However, predicting an infinite number of heatmaps for a continuum of points is not possible, as it would require storing an infinite number of last-layer weights.
With our Neural Localizer Fields, we can harness the well-performing heatmap paradigm and adapt it to predicting keypoints chosen at test time.
 
\paragraph{Parametric fitting to nonparametric predictions}
Fitting a parametric mesh to keypoint predictions is needed for certain applications and for compactness.
If correspondences are known, the classical way is to solve a nonlinear least squares problem as in the original SMPL~\cite{Loper15TOG} paper, but this requires many iterations and is slow.
To speed up and to make the process differentiable, some works train an MLP to predict body parameters from sparse markers~\cite{kolotouros2019cmr,zanfir2021thundr}.
This requires generalizing to all possible poses which is hard. 
By contrast, we propose a solver based on limb rotation estimation via the Kabsch algorithm and linear least-squares shape fitting.
This direct geometry-based algorithm is orders of magnitude faster than a classical optimization procedure, and does not suffer from generalization issues of the MLP methods since it is not learning-based.

\section{Method}
\begin{figure}[t]
\centering
\includegraphics[trim=11mm 3mm 15mm 0mm, clip,width=\linewidth]{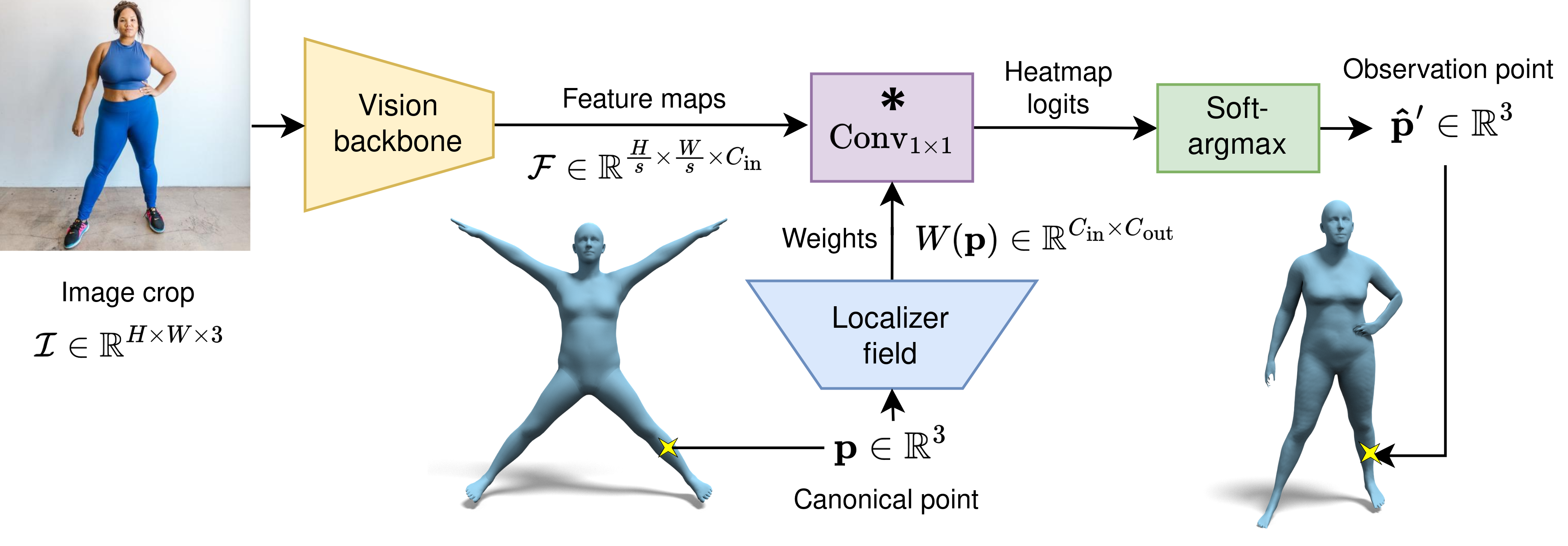}\\
\titledcaption{Overview of NLF}{
Given image features $\mathcal{F}$ and any arbitrarily chosen 3D query point $\p$ within the canonical human volume, we aim to estimate the observation-space 3D point $\p\pr$.
To control which point gets estimated, we dynamically modulate a convolutional layer at the output, to produce heatmaps for the requested point.
We achieve this modulation by predicting the convolutional weights through a neural field.
During training, the points $\p$ can be picked per training example based on whichever points are annotated for it, allowing natural dataset mixing.
At test time, the model can flexibly estimate any surface point and any skeletal joint inside the body volume, as required.}
\label{fig:illust}
\end{figure}

Our method consists of three main parts: (1) a \emph{point localizer network} (PLN) for localizing 3D points from images, (2) a \emph{neural localizer field} (NLF) to control what point the PLN should predict, and (3) an efficient body model fitting algorithm to create a parametric representation for a set of points that were localized nonparametrically.
We metonymically refer to our complete approach as \textit{NLF}.
This modular structure establishes clear interfaces and can enable independent improvements on the parts.
The architecture overview is depicted in \cref{fig:illust}.

\subsection{Point Localizer Network}
We start with describing the basic mechanism of localizing a point in our architecture, which we then generalize to arbitrary points afterward.

Given an RGB image crop $\mathcal{I}\in \brackets{0,1}^{H\times W\times 3}$ of a person, we first extract feature maps $\mathcal{F}\in\R^{h\times w\times C}$ through an off-the-shelf backbone at stride $s=W/w=H/h$.
We then follow the MeTRAbs~\cite{Sarandi21TBIOM,Sarandi23WACV} architecture in skeletal 3D pose estimation, and produce metric-scale truncation-robust volumetric heatmaps $H_\text{3D}\in\R^{h\times w\times D}$, as well as 2D pixel-space heatmaps $H_\text{2D}\in\R^{h\times w}$ through a 1$\times$1 convolutional layer.
The heatmaps are decoded to coordinates via soft-argmax~\cite{Levine16JMLR,Nibali18arXiv,Sun18ECCV}, and the resulting 2D and root-relative 3D points are combined to an absolute camera-space estimate, including estimates for truncated body parts. 
Additionally, we predict an uncertainty map $U\in\R^{h\times w}$, which will be averaged to an uncertainty scalar per point using weights from $H_\text{2D}$.

A classical, joint-based pose estimator would be trained to produce a fixed set of heatmaps, as many as there are joints in the skeleton, and it would have to be trained and tested with the corresponding fixed set of joints.
Taking a functional view, observe that this type of architecture realizes a discrete set of \textbf{localizer functions}, mapping image features to 3D locations as
\begin{equation}
f_i: \R^{h\times w\times C} \rightarrow \R^3, \quad \mathcal{F} \mapsto \phat\pr_i, \quad i\in \brackets{1..J}\\
\end{equation}
with $J$ being the number of joints the model can predict.
In a pose estimator, each $f_i$ mapping is represented through a section of the weight matrix $W$ of the convolutional output head.
Hence, in this paradigm, implementing localizer functions to detect further points requires extending the weight matrix, giving linear growth in the parameters of the output layer.
It is clearly infeasible to store the weights for every possible point we might want to predict.
Furthermore, even though the functions for localizing nearby points of the body have more in common than those detecting faraway points, this spatial structure is lost when representing them as discrete outputs.
In the next part, we describe our method for retrieving the full continuum of localizer functions, allowing predictions for any point, as well as ensuring spatial information sharing across the functions.

\subsection{Neural Localizer Field}
The goal of our work is to go beyond predicting only a predetermined finite set of localizable points in a network, and to instead learn to localize \emph{all} points of the body with one universal model, by learning a whole \emph{localizer field} of functions $\Psi$.
The field $\Psi$ associates a function $f_\p$ to every point $\p$ in the canonical human volume $\Omega_{H} \subset \mathbb{R}^3$:
\begin{equation}
    \Psi: \Omega_{H} \rightarrow \parens{\R^{h\times w\times C} \rightarrow \R^3}, \quad  \p\mapsto \parens{f_\p:\mathcal{F} \mapsto \p\pr},
\end{equation}
where the posed point $\p\pr$ is determined based on image features $\mathcal{F}$ by $f_\p$. 
Although neural fields are typically used to predict points or vectors, here we use them to predict localizer functions $f_\p$.
This way, we represent information on a continuous domain, while appropriately modeling both smooth structure and fine-grained details of a signal.
Specifically, $f_\p(\mathcal{F})$ is implemented as a single-layer convolutional head whose weights are predicted as a function $w(\p)$ of the canonical point $\p$ we want to localize.
The function $w: \Omega_{H} \rightarrow \R^{(C+1)\cdot(D+2)}$ is realized as an MLP, taking three canonical coordinates of $\p$ as input and producing parameters ($\mathbf{W},\mathbf{b}$) for $f_\p$'s convolutional layer to control which point gets predicted (\cf hypernetworks).
Here the $C+1$ is the backbone feature dimensionality plus one bias, and $D+2$ are the total number of output channels which are $D$ depth channels of the volumetric heatmap $H_\text{3D}$, one 2D heatmap $H_\text{2D}$ and one uncertainty map $U$.
In summary, given the image features $\mathcal{F}$ and a canonical query point $\p$, the following computation takes place:
\begin{gather}
(\mathbf{W},\mathbf{b}) := w\parens{\p} \\
(H_\text{2D}, H_\text{3D}, U) := \mathrm{conv}_{1\times1}\parens{\mathcal{F}; \mathbf{W}, \mathbf{b}}\\
\p\pr_{\text{3D, rootrel}} := \operatorcall{soft-argmax}{H_\text{3D}} \quad \quad \p\pr_{\text{2D}} := \operatorcall{soft-argmax}{H_\text{2D}}\\
u := \sum\nolimits_{x,y} U_{x,y} \operatorcall{softmax}{H_\text{2D}}_{x,y} \quad \sigma := \operatorcall{softplus}{u} + \epsilon,
\end{gather}
after which the 3D root-relative and the 2D predictions are fused into a camera-space output $\p\pr$ following~\cite{Sarandi21TBIOM}.
The localizer field's architecture consists of a positional encoding part and an MLP with GELU activation (see the appendix for the details of the MLP layers).

\paragraph{Positional encoding}
To express high-frequency signals, neural fields typically employ positional encodings~\cite{rahaman2019spectral,tancik2020fourier}. 
Since the field's domain is the canonical human volume $\Omega_H$,
we use the eigenbasis $\phi_i$ of the volumetric Laplacian $\Delta$ (the Fourier basis equivalent for bounded volumes), inspired by the use of the Laplace--Beltrami operator in~\cite{neverova2020continuous}.
We then compute the \emph{global point signature} (GPS)~\cite{rustamov2007laplace} 
\begin{equation}
    \gamma\parens{\p} = \brackets{\phi_1(\p)/\sqrt{\lambda_1}, ..., \phi_M(\p)/\sqrt{\lambda_M}}\T \in \R^M,
\end{equation}
a widely used descriptor in shape matching.
So far, this only gives us the eigenfunction values $\phi_i$ on the discrete tetrahedral mesh nodes.
To obtain an approximate interpolant for the entire domain, we distill the eigenfunctions into a small learnable Fourier feature network~\cite{li2021learnable} $h: \R^3 \rightarrow \R^M$ with $h_i\parens{\p}\approx \phi_i\parens{\p}$.
The network $h$ can be finetuned end-to-end.
Details are in the appendix.
We will compare the effectiveness of plain coordinates, global point signature, and learnable Fourier features.

\paragraph{Mathematical formulation for training}
\label{sec:training}
Our formulation allows us to supervise the point-localization task on a continuous domain.
Given an image $\mathcal{I}$ with features $\mathcal{F}$, and the ground-truth continuous warp function $g: \Omega_{H} \rightarrow \R^3$ mapping between the canonical and observation spaces, the loss measures the distance between ground truth $g(\cdot)$ and prediction $f(\cdot; \mathcal{F})$ in function space, through the volume integral
\begin{equation}
\loss{} = \iiint_{\Omega_H} \loss{p}\parens[\big]{f\parens{\p; \mathcal{F}}, g\parens{\p}} \, \mathrm{d}^3\p,
\label{eq:continuous}
\end{equation}
where $\loss{p}(\cdot,\cdot)$ is the pointwise loss comparing prediction $f\parens{\p; \mathcal{F}}=\phat\pr$ to ground truth $g\parens{\p}=\p\pr$ in observation space.
We adopt the Euclidean loss
\begin{equation}
\loss{p}(\phat\pr, \p\pr) = \norm{\phat\pr - \p\pr}
\end{equation}
without squaring, for better outlier-robustness.
(In the appendix, we show how the uncertainty output can be combined into the loss function, as this aspect is not the main focus here.)
This loss is applied to the predictions in the 3D camera coordinate frame, the 3D root-relative frame and the 2D image space as well, with weighting factors.
Naturally, only the 2D loss is applied for training examples that only have 2D labels.

In practice, we often do not have the full volume annotation $g(\p)$ available for all $\p$ as in \eqref{eq:continuous}, and it would be too expensive to evaluate the integral for every image.
Since we train on a mixture of datasets, each training example
$\parens[\big]{\mathcal{I}_i,\{\p\pr_k\}_{k=1}^{K_i}}$ can contain a different annotated point set $\{\p\pr_k\}_{k=1}^{K_i}$ depending on the dataset the example comes from.
Hence, we turn the integral \eqref{eq:continuous} into a discrete sum over the available dataset-specific locations $\p_k \in \Omega_H$.
For training examples with parametric body annotations, we perform Monte Carlo approximation of the integral by random sampling of points.

\paragraph{Zoo of annotations}
Data in HPE contains a wide variety of annotation types including parametric meshes, 3D keypoints, 2D keypoints and DensePose samples.
Thanks to our formulation described above, we can directly consume this data, since the zoo of annotations consists of subsets of points in the canonical volume.
For examples that are annotated with a parametric mesh model, we sample 640 points uniformly over the surface and 384 points from the interior of the human volume.
The sampled points are unique to each example of the batch.
We supervise interior points densely to enable the network to predict new joint sets anywhere in the volume at test time.
To deform interior points, we use scattered data interpolation~\cite{sibson1981brief,bobach2009natural} to propagate SMPL deformations to the volume.
For datasets annotated with 2D or 3D keypoints, the corresponding canonical positions need to be established per joint.
We use an approximate initialization per skeleton type (see appendix), and finetune the precise canonical positions during the main training process.
In case of DensePose annotations, we map the $I,U,V$ labels to the corresponding SMPL surface points in the canonical pose.

\paragraph{Implementation details}
We use EfficientNetV2-S (256 px) and L (384 px)~\cite{Tan21ICML} initialized from \cite{Sarandi23WACV}, and train with AdamW~\cite{Loshchilov19ICLR}, linear warmup and exponential learning rate decay for 300k steps.
Training the S model takes $\sim$2 days on two 40~GB A100 GPUs, while the L takes $\sim$4 days on 8 A100s.

\subsection{Efficient Body Model Fitting in Post-Processing}
\label{sec:fitting}
Above, we introduced our method to nonparametrically localize any human point from an image, in 3D.
While nonparametric prediction has its advantages (\eg, closer fits to the image data), in some cases a parametric representation (joint rotations $\theta$ and shape vector $\beta$) is preferable due to its compactness, and disentanglement of pose and shape.
We therefore develop a fast algorithm to solve for pose and shape parameters that closely reproduce our predicted nonparametric vertices.

Our algorithm alternates between two steps for a small number of iterations ($\sim$3), followed by an adjustment step. We first independently fit global orientations per body part by applying a weighted Kabsch algorithm.
Then, keeping the orientations fixed, we solve for $\beta$ and the translation vector via linear least squares, since SMPL's vertex positions are linear in the shape parameters.
Finally, we refine the part rotations in one pass along the kinematic tree, anchoring the rotations at the parent joint.
For improved efficiency, it is possible to consider only a subset of the vertices when fitting.
Further, given that our model estimates per-point uncertainties, these can be used in a weighted variant of the above algorithm.
Splitting the shape estimation task into a nonparametric point localization and a fitting step has the further benefit that it is easy to share the body shape parameters $\beta$ for multiple observations of the same person (multi-view and temporal use cases) during the least-squares regression.
More details and the pseudocode of the algorithm are given in the appendix.

\section{Experiments}
We extensively evaluate our method on a variety of benchmarks: 3DPW~\cite{VonMarcard18ECCV} and EMDB~\cite{kaufmann2023emdb} for SMPL body, AGORA~\cite{Patel21CVPR} and EHF~\cite{Pavlakos19CVPR} for SMPL-X, SSP-3D~\cite{STRAPS2020BMVC} for SMPL focusing on body shape, as well as Human3.6M~\cite{Ionescu14PAMI}, MPI-INF-3DHP~\cite{Mehta17TDV} and MuPoTS-3D~\cite{Mehta18TDV} for 3D skeletons.

Our goal is to enable large-scale multi-dataset training from heterogeneous annotation sources in order to train strong models.
To show that our formulation is effective at this, we assemble a large meta-dataset for training, extending PosePile\footnote{\url{https://github.com/isarandi/posepile}}, introduced in~\cite{Sarandi23WACV}.
We provide dataset details in the appendix.
For parametric meshes we include the datasets \cite{Patel21CVPR,Black_CVPR_2023,Bhatnagar22CVPR,Huang:CVPR:2022,Kocabas21ICCV,tripathi2023ipman,fan2023arctic,huang2022intercap,Cheng22Arxiv,zhang2022egobody,yin2023hi4d,cai2022humman,yang2023synbody,tao2021function4d,peng2021neural,dfaust:CVPR:2017,Varol17CVPR}, with points sampled directly from each dataset's native body model (one of SMPL, SMPL-X or SMPL-H, neutral or gendered, as provided) without a need for any conversions.
For 3D skeleton annotations, we use \cite{Ionescu14PAMI,Mehta17TDV,Mehta18TDV,Yu20CVPR,Zhang20CVPR,Joo17PAMI,Li21ICCV,Nibali21IVC,Wang19Arxiv,Pumarola18CVPR,Hu19CVPR,GhorbaniPLOSOne21,Zhang17IVC,Aa11ICCVW,Ofli13WACV,Trumble19BMVC,Fabbri18ECCV,BenShabat21WACV,Fieraru21CVPR,Fieraru20CVPR,Fieraru21AAAI,Bazavan21Arxiv,khirodkar2023egohumans,2023dnarendering}. 
Several of these have custom skeletons that are nontrivial to convert to the SMPL family, preventing existing 3D HPS methods from exploiting these rich data sources.
For 2D keypoint annotations we use \cite{Insafutdinov16ECCV,jin2020whole,Vendrow22Arxiv,Andriluka18CVPR,wu2017ai,alphapose} and further use DensePose datasets~\cite{Neverova18ECCV,Guler18CVPR} which provide pairs of SMPL surface points and corresponding pixel locations.
Integrating all these datasets would be a great challenge if we had to choose one specific representation.
Instead, we can easily train using the strategy described in \cref{sec:training}, we simply map each annotation convention to the canonical human volume and learn a continuous localizer field.

We use standard metrics including per-joint error (MPJPE), per-vertex error (MVE), orientation angle error (MPJAE) and their Procrustes variants (\enquote{P-}).
Metrics are explained in detail in the appendix.

\begin{table}[tp]
\label{tab:positional-encoding}
\setlength{\tabcolsep}{2.5pt}%
\titledcaption{Ablation of positional encodings}{Input representation matters in NLF.
The fixed Laplacian-derived global point signature helps most metrics, but is outperformed by learnable Fourier features. 
Best results are achieved by initializing the learnable Fourier features to the GPS.}
\begin{tabularx}{\linewidth}{lcccc}
\toprule
& SSP3D mIoU$\uparrow$ & 3DPW P-MVE$\downarrow$ & EMDB P-MVE$\downarrow$ & 3DPW MPJAE$\downarrow$ \\
\midrule
Plain XYZ coordinates & 0.777 & 54.0 & 57.7 & 16.9 \\
Global point signature (GPS) & 0.789 & 54.2 & 57.5 & 15.8 \\
Learnable Fourier features & 0.804 & \B52.8 & 57.0 & 15.4 \\
Learnable Fourier (GPS init.) & \B0.812 & \B52.8 & \B56.4 & \B 15.3 \\
\bottomrule
\end{tabularx}
\end{table}%

\paragraph{Pose prediction}
NLF outperforms all baselines on \textbf{EMDB} (\cref{tab:emdb}), which has challenging outdoor sequences.
The improvement is drastic, we get an MPJPE of 68.4~mm compared to the second-best (98.0~mm).
Remarkably, we outperform the temporal method WHAM (79.7~mm) although we work frame-by-frame.
\textbf{3DPW} results confirm the same (\cref{tab:3dpw14}).
(For reference, the standard error of the mean under normal assumptions for 3DPW MPJPE is 0.06~mm.)
We attribute this performance to the fact that we can natively train on many sources without error-prone conversions.
(For better comparability to temporal methods, we also provide NLF results using a (non-learned) 5-frame temporal smoothing filter.)
On \textbf{AGORA} (\cref{tab:agora_test}), we outperform all prior works in SMPL-X prediction.
Even though we do not specially target facial and hand keypoints, we obtain the second best scores for these.
We also achieve state-of-the-art body results on \textbf{EHF} (\cref{tab:ehf}). The higher hand error is due to few detailed hands in the training data.
We obtain excellent performance on the skeleton estimation benchmarks \textbf{Human3.6M, MPI-INF-3DHP and MuPoTS-3D} as well (\cref{tab:h36m}).
This evaluation is easy with NLF, as we can pick at test time \emph{which points of the body to predict}.

\begin{figure}[tp]
\centering
\includegraphics[width=\linewidth]{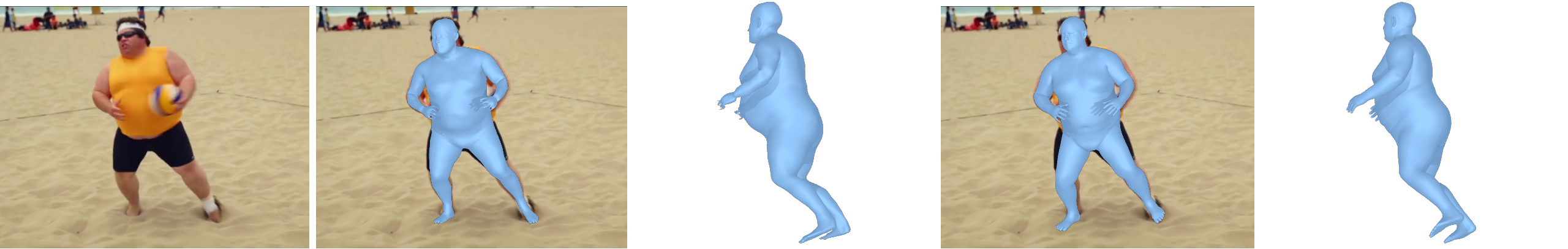}\\
\titledcaption{Qualitative result on SSP-3D}{\emph{left:} NLF's nonparametric output (front and side view), \emph{right:} result of our proposed fast SMPL fitting algorithm (front and side). Our nonparametric prediction already has high quality, allowing us to use a simple and efficient fitting algorithm to obtain body model parameters that faithfully represent the nonparametric output.}
\label{fig:ssp3d}
\end{figure}

\paragraph{Shape prediction}
Mesh recovery methods are often biased towards an average shape.
Keypoint-based methods are pixel-aligned but ignore shape, and SMPL parametric regressors tend to produce subpar image alignment. 
By contrast, since we model the full continuum of points, we can estimate shape more accurately while being pixel-aligned.
This is reflected on \textbf{SSP-3D} (\cref{tab:ssp3d}), a dataset specifically designed to evaluate shape estimation, where we again improve over all baselines, obtaining a low error of 10.0~mm.
Notably, we outperform works, such as ShapeBoost~\cite{Bian24AAAI}, which are specialized to shape estimation and have not been tested on general pose estimation benchmarks.
\Cref{fig:ssp3d} shows a qualitative example.
Following the protocol in \cite{sengupta2021probabilistic}, we also evaluate combined shape estimation from multiple (max. 5) images of the same person, by shared estimation of the $\beta$ shape parameters during body model fitting.
This reduces the error to 9.6~mm.
Our NLF achieves state-of-the-art shape estimation while \emph{simultaneously} obtaining SOTA results on the most challenging pose benchmarks such as EMDB, using the same model weights.

\paragraph{Positional encoding} 
To assess the impact of positional encoding on shape fidelity, we evaluate the mIoU silhouette overlap measure on the shape-diverse SSP-3D, the Procrustes-aligned vertex error on 3DPW and EMDB, as well as the orientation error on 3DPW after SMPL-fitting using the small backbone.
As seen in \cref{tab:positional-encoding}, the use of the Laplacian-based global point signature (GPS) encoding leads to a better mIoU of $0.789$ compared to the $0.777$ achieved using plain $X,Y,Z$ coordinates.
We attribute this to better representation of fine-grained geometry.
Learnable Fourier features obtain even better results than GPS, but best is the combination where we use the GPS as initialization for the learnable Fourier feature network.
Do note however that NLF achieves good performance even with plain coordinates.

\paragraph{Uncertainty estimation} 
We provide an ablation for uncertainty predictions in the appendix.

\newcommand{\gr}{\itshape\color{darkgray}}
\begin{table}[p]
\centering
\setlength{\tabcolsep}{3pt}
\titledcaption{Results on 3DPW (14 joints)}{* denotes \emph{temporal} (multi-frame) method}
\resizebox{\textwidth}{!}{
\small
\begin{tabularx}{\linewidth}{@{}lcccc|cccc@{}}
\toprule
& \multicolumn{4}{c|}{Trained without 3DPW-train} & \multicolumn{4}{c}{Trained with 3DPW-train} \\
\cmidrule(lr){2-5} \cmidrule(l){6-9}
Method & MPJPE & P-MPJPE & MVE & P-MVE & MPJPE & P-MPJPE & MVE & P-MVE \\
\midrule
PyMAF~\cite{pymaf2021} & 78.0 & 47.1 & 91.3 &  & 74.2 & 45.3 & 87.0 &  \\
SMPLer-X-H32~\cite{cai2023smplerx} & 75.0 & 50.6 &  &  & 71.7 & 48.0 &  &  \\
BEDLAM-CLIFF~\cite{Black_CVPR_2023} & 72.0 & 46.6 & 85.0 &  & 66.9 & 43.0 & 78.5 &  \\
HybrIK~\cite{Li21CVPR}  & 71.6 & 41.8 & 82.3 &  &  &  &  &  \\
HMR 2.0a~\cite{goel2023humans} & 70.0 & 44.5 &  &  &  &  &  &  \\
Multi-HMR~\cite{baradel2024multi} & 69.5 & 46.9 & 88.8 & & 61.4 & 41.7 & 75.9 & \\
\emph{WHAM-B (ViT, w/ BEDLAM)}{\B*}~\cite{shin2023wham} &  &  &  &  & \gr56.9 & \gr35.7 & \gr67.4 \\
\midrule
NLF-S & 60.9 & 38.5 & 73.3 & 52.8 & 55.6 & 35.9 & 67.0 & 48.9 \\
NLF-S +fit & 60.8 & 37.9 & 72.2 & 51.4 & 56.6 & 35.7 & 66.7 & 47.9 \\
NLF-L & 60.3 & 37.3 & 71.4 & 50.2 & \B54.1 & 33.7 & \B63.7 & 45.3 \\
NLF-L +fit & \B59.0 & \B36.5 & \B69.7 & \B48.8 & 54.9 & \B33.6 & \B63.7 & \B44.5 \\
NLF-L +fit \emph{+smooth}{\B*} & \gr \B57.2 &\gr \B35.4 & \gr\B67.8 & \gr\B47.7 & \gr\B53.2 & \gr\B32.6 & \gr\B62.1 & \gr\B43.5 \\
\bottomrule
\end{tabularx}
}
\label{tab:3dpw14}
\end{table}

\newcolumntype{R}[2]{%
    >{\adjustbox{angle=#1,lap=\width-(#2)}\bgroup}%
    l%
    <{\egroup}%
}
\newcommand*\rot{\multicolumn{1}{R{45}{1em}}}

\begin{table}[tp]
\centering
\begin{minipage}{0.55\textwidth}
        \centering
\titledcaption{Results on EMDB1 (24j)}{*\emph{temporal} method}
\setlength{\tabcolsep}{1.5pt}
\resizebox{1.00\linewidth}{!}{
\begin{tabular}{lcc|cc|cc}
\toprule
Method & \rot{MPJPE} $\downarrow$ & \rot{P-MPJPE}  $\downarrow$ & \rot{MVE}  $\downarrow$ & \rot{P-MVE}  $\downarrow$ & \rot{MPJAE}  $\downarrow$ & \rot{P-MPJAE}  $\downarrow$ \\
\midrule
PyMAF \cite{pymaf2021} & 131.1  & \z82.9  & 160.0  & \z98.1  & \z28.5  & \z25.7   \\
LGD \cite{Song20ECCV} & 115.8  & \z81.1  & 140.6  & \z95.7  & \z25.2  & \z25.6  \\
ROMP \cite{ROMP} & 112.7  & \z75.2  & 134.9  & \z90.6  & \z26.6  & \z24.0  \\
PARE \cite{Kocabas21ICCVb} & 113.9  & \z72.2  & 133.2  & \z85.4  & \z24.7  & \z22.4 \\
GLAMR$^*$ \cite{Yuan22CVPR} & 107.8  & \z71.0  & 128.2  & \z85.5  & \z25.5  & \z23.5  \\
FastMETRO-L \cite{cho2022FastMETRO} & 115.0  & \z72.7  & 133.6  & \z86.0  & \z25.1  & \z22.9 \\
CLIFF \cite{li2022cliff} & {103.1}  & \z68.8  & 122.9  & \z81.3  & {\z23.1}  & {\z21.6}  \\
HybrIK \cite{Li21CVPR} & {103.0}  & {\z65.6}  & {122.2}  & {\z80.4}  & {\z24.5}  & \z23.1   \\
HMR 2.0~\cite{goel2023humans} &  \z98.0& \z60.6& 120.3& & & \\
BEDLAM-CLIFF~\cite{Black_CVPR_2023} &  \z98.0 &\z60.6& 111.6& & & \\
\emph{WHAM-B (ViT)}{\B*}~\cite{shin2023wham} & \z\gr79.7 & \z\gr50.4 & \z\gr94.4 & & & \\ 
\midrule
NLF-S & \z74.5 & \z44.9 & \z87.8 & \z56.4 & -- & -- \\
NLF-S +fit & \z72.0 & \z44.6 & \z85.2 & \z55.4 & \z17.1 & \z16.4 \\
NLF-L & \z69.6 & \z41.2 & \z82.4 & \z52.0 & -- & -- \\
NLF-L +fit & \B\z68.4 & \B\z40.9 & \B\z80.6 & \B\z51.1 & \B\z16.1 & \B\z15.4 \\
NLF-L +fit \emph{+smooth}{\B*} & \z\gr\B66.7 & \z\gr\B39.9 & \z\gr\B78.7 & \z\gr\B50.0 & \z\gr\B16.0 & \z\gr\B15.2 \\
\bottomrule
\end{tabular}
}
\label{tab:emdb}
\end{minipage}
    \hfill
    \begin{minipage}{0.42\textwidth}
\titledcaption{Results on SSP-3D}{
PVE-T-SC is the per-vertex error in T-pose with scale correction, mIoU measures silhouette overlap.
* means estimating body shape from multiple images of a person (max. 5).}
\label{tab:ssp3d}
\resizebox{1.00\linewidth}{!}{
\begin{tabular}{@{}lcc@{}}
\toprule
Method & PVE-T-SC$\downarrow$ & mIoU$\uparrow$  \\
\midrule
HMR~\cite{Kanazawa18CVPR}    & 22.9 & 0.69 \\
SPIN~\cite{kolotouros2019learning}  & 22.2 & 0.70 \\
SHAPY~\cite{Shapy:CVPR:2022}  & 19.2 & 0.71 \\
SoY~\cite{sarkar2023shape}    & 15.8 & 0.76 \\
STRAPS~\cite{STRAPS2020BMVC} & 15.9 & 0.80 \\
Sengupta \etal \cite{sengupta2021probabilistic} & 15.2 & -- \\
Sengupta \etal* \cite{sengupta2021probabilistic} & \gr 13.3 & -- \\
Sengupta \etal \cite{sengupta2021hierarchical} & 13.6 & -- \\
LASOR \cite{yang2022lasor} & 14.5 & 0.67 \\
HuManiFlow~\cite{sengupta2023humaniflow} & 13.5 & -- \\
KBody~\cite{zioulis2023kbody} & 25.6 & 0.80 \\
ShapeBoost~\cite{Bian24AAAI} & 11.4 & -- \\
\midrule
NLF-S +fit & \B10.0 & \B0.85 \\
NLF-S +fit (\emph{shared} $\beta$)* & \z\gr\B9.6 & \gr\B0.85 \\
NLF-L +fit & 11.1 & 0.83 \\
\bottomrule
\end{tabular}
}
\end{minipage}
\end{table}

\begin{table}[tp]
  \centering
  \titledcaption{Results on AGORA-test (SMPL-X)}{Models are fine-tuned on AGORA.%
  }
  \label{tab:agora_test}
  \resizebox{\textwidth}{!}{
  \setlength{\tabcolsep}{1.5pt}
  \begin{tabular}{lccccc|ccccc|ccccc}
    \toprule
    & 
    \multicolumn{2}{c}{NMVE $\downarrow$} &
    \multicolumn{2}{c}{NMJE $\downarrow$} &
    \multicolumn{5}{c}{MVE $\downarrow$} &
    \multicolumn{5}{c}{MPJPE $\downarrow$} \\
    \cmidrule(lr){2-3} \cmidrule(lr){4-5} \cmidrule(lr){6-10} \cmidrule(lr){11-15}  
    Method &
    All & Body &
    All & Body &
    All & Body & Face & LHan & RHan &
    All & Body & Face & LHan & RHan 
    \\
    \midrule
    Hand4Whole~\cite{Moon_2022_CVPRW_Hand4Whole}& 
    144.1 & \z96.0 & 141.1 & \z92.7 & 135.5 & \z90.2 & \z41.6 & \z46.3 & \z48.1 & 132.6 & \z87.1 & \z46.1 & \z44.3 & \z46.2 \\
    BEDLAM~\cite{Black_CVPR_2023} & 
    142.2 & 102.1 & 141.0 & 101.8 & 103.8 & \z74.5 & \B \z23.1 & \B \z31.7 & \B \z33.2 & 102.9 & \z74.3 & \B \z24.7 & \B \z29.9 & \B\z31.3 \\
    PyMAF-X~\cite{pymafx2023} &
    141.2 & \z94.4 & 140.0 & \z93.5 & 125.7 & \z84.0 & \z35.0 & \z44.6 & \z45.6 & 124.6 & \z83.2 & \z37.9 & \z42.5 & \z43.7 \\
    OSX~\cite{lin2023one}&
    130.6 & \z85.3 & 127.6 & \z83.3 & 122.8 & \z80.2 & \z36.2 & \z45.4 & \z46.1 & 119.9 & \z78.3 & \z37.9 & \z43.0 & \z43.9 \\
    HybrIK-X~\cite{li2023hybrikx} & 
    120.5 & \z73.7 & 115.7 & \z72.3 & 112.1 & \z68.5 & \z37.0 & \z46.7 & \z47.0 & 107.6 & \z67.2 & \z38.5 & \z41.2 & \z41.4 \\
    SMPLer-X~\cite{cai2023smplerx} & 
    107.2 & \z68.3  &104.1 & \z66.3 & \z99.7 & \z63.5  &\z29.9  &\z39.1 & \z39.5  &\z96.8  &\z61.7  &\z31.4 & \z36.7  &\z37.2 \\
    Multi-HMR~\cite{baradel2024multi} &
    102.0 & \z63.4 & 101.8 & \z64.1 & \z95.9 & \z59.6 & \z27.7 & \z40.2 & \z40.9 & \z95.7 & \z60.3 & \z29.2 & \z38.1 & \z39.0 \\
    \midrule
    NLF-S &
    108.6 & \z67.9 & 106.3& \z67.4 & 102.1 & \z63.8 & \z29.5 & 	\z42.7 & \z42.8 & \z99.9 & \z63.4 & \z31.7 & 	\z38.7 & \z39.2\\
    NLF-L &
    \z\B{98.6} & \z\B{62.1} & \z\B{96.6} & \z\B{61.9} & \B\z92.7 & \z\B{58.4} & \z27.0 & \z{37.9} & \z{38.1} & \B\z90.8 & \z\B{58.2} & \z28.5 & \z{34.4} & \z{34.9} \\
    \bottomrule
  \end{tabular}}
\end{table}
\begin{table}[tp]
\centering
\setlength{\tabcolsep}{3pt}
\titledcaption{Results on skeleton estimation benchmarks}{}
\begin{tabularx}{\linewidth}{@{}lcc|ccc|c@{}}
\toprule
Method & \multicolumn{2}{c}{Human3.6M} & \multicolumn{3}{c}{MPI-INF-3DHP} &MuPoTS-3D \\
    \cmidrule(lr){2-3} \cmidrule(lr){4-6} \cmidrule(lr){7-7}
 & MPJPE$\downarrow$ & P-MPJPE$\downarrow$ & PCK$\uparrow$ & AUC$\uparrow$ & MPJPE$\downarrow$ & PCK-detected$\uparrow$ \\
\midrule 
MeTRAbs-ACAE-S~\cite{Sarandi23WACV} & 40.2 & 31.1 & 96.3 & 58.7 & 57.9 & 94.7\\
MeTRAbs-ACAE-L~\cite{Sarandi23WACV} & \B36.5 & \B27.8 & 97.1 & 60.1 & 55.4 & 95.4 \\
\midrule
NLF-S & 40.4 & 30.6 & 96.6 & 57.9 & 59.9 & 94.7 \\
NLF-L & 39.7 & 28.5 & \B97.5 & \B61.0 & \B54.9 & \B95.5 \\
\bottomrule
\end{tabularx}
\label{tab:h36m}
\end{table}
\begin{table}[tp]
  \centering
  \titledcaption{Results on EHF (SMPL-X)}{}
  \label{tab:ehf}
  \setlength{\tabcolsep}{3.5pt}
\begin{tabular}{lccccccccc}
\toprule
Method & \multicolumn{3}{c}{MVE} & \multicolumn{4}{c}{P-MVE} & \multicolumn{2}{c}{P-MPJPE} \\
\cmidrule(lr){2-4} \cmidrule(lr){5-8} \cmidrule(lr){9-10}
 & Full-body & Hands & Face & Full-body & Body & Hands & Face & Body (14j) & Hands \\
\midrule
HybrIK-X~\cite{li2023hybrikx} & 121.4 & 52.1 & 41.9 & 59.8 & 50.0 & 17.6 & 8.1 & 60.8 & 17.9 \\
OSX~\cite{lin2023one} & 70.8 & 53.7 & 26.4 & 48.7 & -- & 15.9 & 6.0 & -- & -- \\
PyMAF-X~\cite{pymafx2023} & 64.9 & 29.7 & 19.7 & 50.2  & 44.8 & \B10.2  & 5.5 & 52.8 & \B 10.3 \\
SMPLer-X~\cite{cai2023smplerx} & 62.4 & 47.1 & 17.0 & 37.1 & -- & 14.1 & \B 5.0 & -- & -- \\
Multi-HMR~\cite{baradel2024multi} & 42.0& \B 28.9 & 18.0 &  28.2 & -- & 10.8 & 5.3 & -- & -- \\
\midrule
NLF-S & 39.7 & 49.7 & 15.2 & 30.9 & 30.4 & 24.5 & 7.4 & 32.8 & 28.2 \\
NLF-S +fit & 40.0 & 49.7 & 15.4 & 30.4 & 29.7 & 24.0 & 6.5 & 32.0 & 25.4 \\
NLF-L & \B 36.3 & 43.2 & \B 13.5 & \B 25.8 & 24.8 & 21.3 & 6.7 & 26.3 & 25.3 \\
NLF-L +fit & 36.4 & 43.0 & 13.9 & 26.0 & \B 24.4 & 20.7 & 6.3 & \B 26.1 & 21.8 \\
    \bottomrule
  \end{tabular}
\end{table}

\begin{figure}[tp]
\centering
\includegraphics[width=\linewidth]{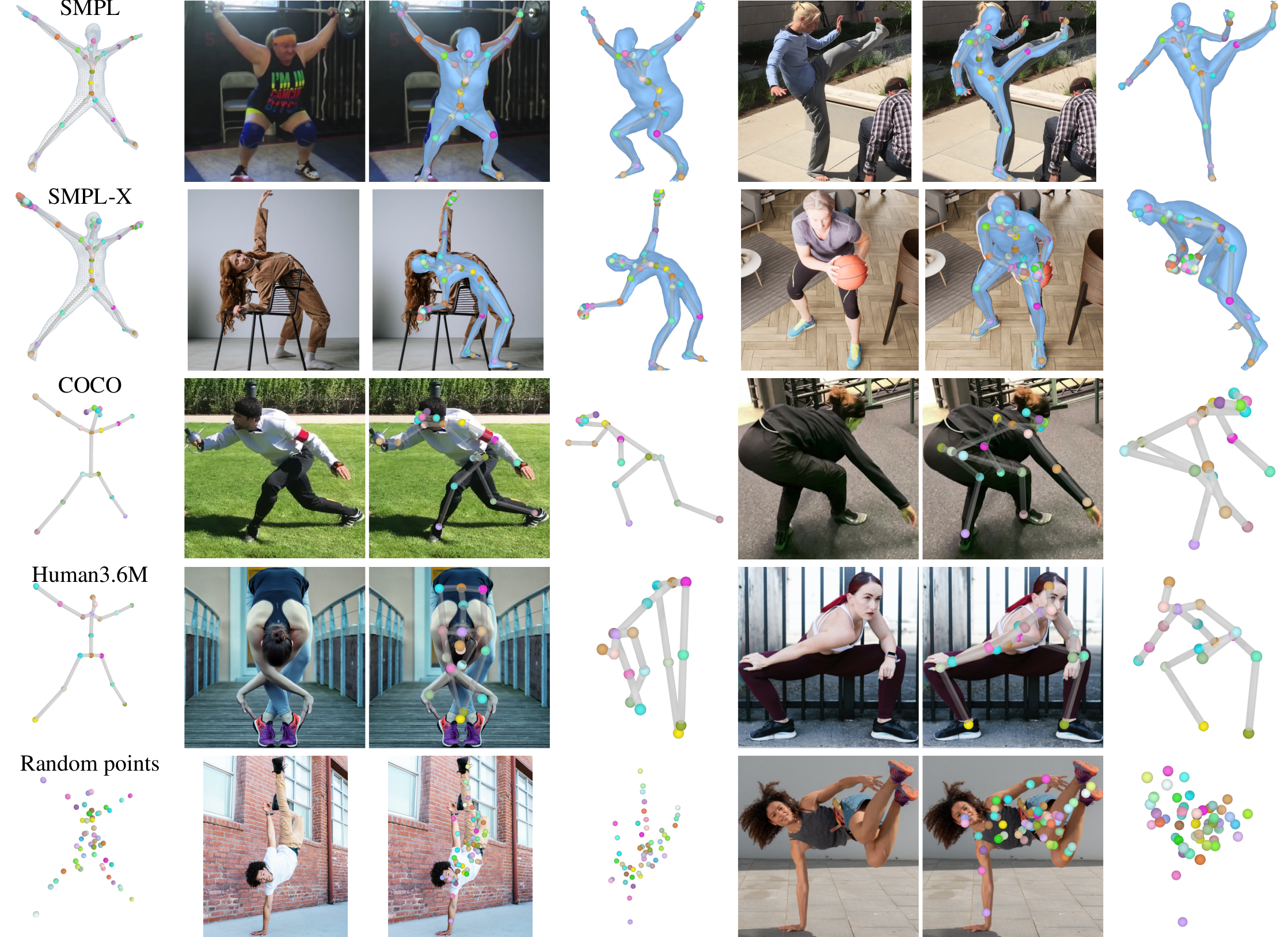}\\
\titledcaption{Customizable point localization}{%
By selecting points $\p$ in the continous canonical space, we can predict any landmark set both at training and test time.
The first column depicts the query points we estimate: SMPL(-X) joints and vertices, COCO joints, Human3.6M joints, and arbitrary points sampled within the human volume.
The fourth and seventh column show rotated views.}
\label{fig:results}
\end{figure}
\begin{table}[tp]
\centering
\small
\setlength{\tabcolsep}{3pt}
\titledcaption{Training data ablation}{Using all datasets yields the best results compared to only the real subset or only the synthetic subset, showing that NLF benefits from both kinds of data and that more data improves results. (Here we trained for 100k steps as opposed to 300k for the main evaluations.)}
\begin{tabular}{lccccccc}
\toprule
 & SSP-3D & H36M & 3DHP & 3DPW & 3DPW & EMDB & EMDB \\
Method / data & PVE-T-SC$\downarrow$ & MPJPE$\downarrow$ & PCK$\uparrow$ & P-MPJPE$\downarrow$ & P-MVE$\downarrow$ & P-MPJPE$\downarrow$ & P-MVE$\downarrow$ \\
\midrule
NLF-S, real datasets only & 11.9 & 43.0 & 96.0 & 40.4 & 55.5 & 47.8 & 59.3 \\
NLF-S, synthetic datasets only & 10.0 & 48.0 & 95.4 & 39.2 & 53.5 & \bfseries45.7 & 57.9 \\
NLF-S, all datasets & \bfseries9.9 & \bfseries41.6 & \bfseries96.3 & \bfseries38.4 & \bfseries53.1 & 45.8 & \bfseries57.2 \\
\bottomrule
\end{tabular}
\label{tab:data_ablation}
\end{table}
\paragraph{Effect of datasets} Assessing the contribution of each individual dataset is computationally infeasible, 
nonetheless, to validate that NLF benefits from diverse supervision, we partition the data into real and synthetic subsets.
As \cref{tab:data_ablation} shows, best results are achieved when all datasets are used in training.
\paragraph{Real-time inference} NLF-S has a batched throughput of 410 fps and unbatched throughput of 79 fps on an Nvidia RTX 3090 GPU. For NLF-L these are 109 fps and 41 fps respectively.
\paragraph{Qualitative results} To demonstrate the versatility of our method in localizing any point of the human body, we show qualitative results in \cref{fig:results} according to different representations.
Qualitative results with the final fitted SMPL are given in the appendix.

\paragraph{Efficient body model fitting}
On both 3DPW and EMDB, fitting SMPL to the nonparametric estimation (\cref{sec:fitting}) generally reduces the error, but the difference is small, indicating that the original predictions already have high quality.
To measure the fitting speed, we consider the task of fitting SMPL-X to vertices derived from SMPL (model transfer) using sample data from the SMPL-X website (33 meshes).
The official code for the transfer takes 33 minutes with mean vertex error 5.0~mm.
Our algorithm takes 28.4~ms with error 7.8~mm, \ie orders of magnitude faster, and error increase is negligible relative to the typical prediction error of centimeters.

\paragraph{Limitations}
NLF works frame-by-frame without temporal cues, and estimates each person independently, so highly-overlapping people remain challenging.
Further, our predictions may have self-intersections and the absolute distance may be inaccurate due to depth/scale ambiguity.

\paragraph{Broader social impact} Human pose and shape estimation can help advance assistive technologies, human--robot interaction or interactive entertainment. However, like other vision methods, a potential for misuse exists in \eg, illegitimate surveillance. We will release models for research only.

\section{Conclusion}
We proposed \emph{Neural Localizer Fields} to upgrade 3D human pose estimators from fixed joint sets to predicting any point of the body surface and volume.
We store point-localizer functions in a neural field, as opposed to storing explicit convolutional weights for every joint the network can predict.
This enables us to train a large-scale generalist pose and shape estimation model, without any effort on relabeling all data to one format.
Instead, we can supervise our model with any points that happen to be annotated for particular training examples.
Besides this training advantage, we can choose at runtime which points in the volume to predict, allowing us to output any desired format.
To make our output even more useful in downstream applications, we proposed an efficient points-to-parameters fitting algorithm to obtain SMPL parameters.
Our model trained on several datasets with different formats achieves state-of-the-art performance with significant improvement over baselines on several benchmarks.
Our model exhibits good performance on outdoor, indoor and synthetic data, regardless of the skeleton/mesh format used for different benchmarks.
With the recent trend of more datasets from different sources being available, NLF paves the way to leverage these elegantly and efficiently.

\begin{ack}
This work was supported by the German Federal Ministry of Education and Research (BMBF): Tübingen AI Center, FKZ: 01IS18039A.
This work is funded by the Deutsche Forschungsgemeinschaft (DFG, German Research Foundation) -- 409792180 (Emmy Noether Programme, project: Real Virtual Humans).
GPM is a member of the Machine Learning Cluster of Excellence, EXC number 2064/1 -- Project number 390727645.
The project was made possible by funding from the Carl Zeiss Foundation.
\end{ack}

{
\small
\setlength{\bibsep}{1.1ex plus 1ex}
\bibliographystyle{abbrvnat2}
\bibliography{abbrev_short,ms}
}

\clearpage
\appendix
\section*{Appendix: Neural Localizer Fields for Continuous 3D Human Pose and Shape Estimation}

In this appendix, we provide further technical details on

\begin{itemize}
\item the experimental setup (datasets, training configuration),
\item uncertainty estimation evaluation,
\item our proposed efficient parametric body model fitting algorithm (pseudocode, convergence analysis, fitting to subsets of vertices, uncertainty-based weighted fitting),
\item our use of positional encodings,
\item the neural field MLP architecture,
\item qualitative results.
\end{itemize}

For video results, including visual comparison to prior work, we refer to our supplementary video. This video includes a demonstration of continuous querying in the canonical space.

\section{Mixture Dataset Description}
The design goal of our work is to naturally mix data sources using different annotation formats.
To show this capability, we perform mixed-batch training with batch size 160, where every batch contains examples with parametric annotations, examples with 3D keypoint annotations, examples with 2D keypoint annotations and examples with DensePose annotations.
In the following, we describe the specific datasets we use for each of these types of annotations.

\subsection{Parametric Annotations}
We use the following datasets with parametric annotations, grouped by the body model:
\begin{itemize}
\item SMPL-X: AGORA~\cite{Patel21CVPR}, BEDLAM~\cite{Black_CVPR_2023},  RICH~\cite{Huang:CVPR:2022}, MOYO~\cite{tripathi2023ipman}, ARCTIC~\cite{fan2023arctic}, InterCap~\cite{huang2022intercap}, GeneBody~\cite{Cheng22Arxiv}, EgoBody~\cite{zhang2022egobody}
\item SMPL: SPEC~\cite{Kocabas21ICCV}, Hi4D~\cite{yin2023hi4d}, HuMMan~\cite{cai2022humman}, SynBody-NeRF~\cite{yang2023synbody}, THuman2.0~\cite{tao2021function4d}, ZJU-Mocap~\cite{peng2021neural}, DFAUST~\cite{dfaust:CVPR:2017}, and SURREAL~\cite{Varol17CVPR}.
\item SMPL+H: BEHAVE~\cite{Bhatnagar22CVPR}
\end{itemize}

\paragraph{Rendering} DFAUST does not provide RGB images but contains diverse shapes that are useful for training, so we render 26K synthetic images with Blender, using randomized SMPLitex~\cite{casas2023smplitex} textures, cameras and lights.
Similarly, since Hi4D contains high-quality fits with good image-alignment, as well as high-quality textured meshes, we also render 100K synthetic images of these meshes with random camera angles and lights.

\subsubsection{Shape Accuracy Considerations}
Datasets have varying levels of shape accuracy, depending on the annotation pipeline used to create them, which means that we need to be careful about what we can treat as \enquote{ground truth} for training purposes.

Several datasets include annotations in SMPL(-X) format, but these were obtained by fitting \emph{only to sparse skeletal body joint locations}, which were in turn triangulated from 2D skeletons (obtained e.g. from OpenPose).
Therefore, the surface shape information in these fits does not line up well with the image, and the body shapes are biased towards the mean shape.
To avoid similarly biasing our \emph{model} to the mean shape, we only treat the joint annotations as valid in these datasets but do not use the surface.
This applies to GeneBody, HuMMan and EgoBody.

Some datasets provide both the SMPL(-X) fit and raw body joints, either from triangulation or motion capture systems.
We observed that the raw body joints tend to line up better with the image, so we ignore the SMPL(-X) fits in these cases.
These datasets are: AIST-Dance++, DNA-Rendering, HUMBI and EgoHumans (these provide COCO-style keypoints besides the body fits) and BML-MoVi (which provides more than 80 markers on the body surface, as well as body joints -- these exhibit better image-alignment than the provided SMPL fits).

\subsection{3D Skeleton Annotations}
We use skeleton annotations (non-SMPL-based) from the following dataset sources: Human3.6M~\cite{Ionescu14PAMI,Ionescu11ICCV}, MPI-INF-3DHP~\cite{Mehta17TDV}, MuCo-3DHP~\cite{Mehta17TDV}, HUMBI~\cite{Yu20CVPR}, 3DOH~\cite{Zhang20CVPR}, CMU-Panoptic~\cite{Joo17PAMI}, AIST-Dance++~\cite{Li21ICCV}, ASPset510~\cite{Nibali21IVC}, GPA~\cite{Wang19Arxiv}, 3DPeople~\cite{Pumarola18CVPR}, SAIL-VOS~\cite{Hu19CVPR}, BML-MoVi~\cite{GhorbaniPLOSOne21}, MADS~\cite{Zhang17IVC}, UMPM\cite{Aa11ICCVW}, Berkeley MHAD~\cite{Ofli13WACV}, TotalCapture~\cite{Trumble19BMVC}, JTA~\cite{Fabbri18ECCV}, IKEA-ASM~\cite{BenShabat21WACV}, Fit3D~\cite{Fieraru21CVPR}, CHI3D~\cite{Fieraru20CVPR}, HumanSC3D~\cite{Fieraru21AAAI}, HSPACE~\cite{Bazavan21Arxiv}, EgoHumans~\cite{khirodkar2023egohumans}, and DNA-Rendering~\cite{2023dnarendering}.

Some of these datasets generated their annotations via multi-view triangulation of 2D pose estimation results, others used marker-based motion capture, and yet others are rendered through graphics and are annotated according to the skeleton of the underlying 3D assets, which may deviate from the SMPL skeleton.

\subsection{2D Annotations}
We are able to use training examples that only provide 2D keypoint information -- in this case, the loss is computed solely by comparing the projection of our model's output to the ground-truth pixel positions.
We use the following 2D keypoint datasets:
MPII~\cite{Insafutdinov16ECCV}, COCO-WholeBody~\cite{Lin14ECCV,jin2020whole,xu2022zoomnas}, JRDB-Pose~\cite{MartinMartin21PAMI,Vendrow22Arxiv}, PoseTrack~\cite{Andriluka18CVPR}, AI Challenger~\cite{wu2017ai} and Halpe~\cite{alphapose}.

We also use the DensePose annotations from DensePose-COCO~\cite{Guler18CVPR} and DensePose-PoseTrack~\cite{Neverova18ECCV}.

\section{Evaluation Protocol}
On 3DPW, we evaluate 14 joints, obtained through the same Human3.6M-style joint regressor that all prior works use.
We only report evaluation results on the test set section of 3DPW.

\paragraph{Bounding boxes}
For testing, some datasets provide bounding boxes (\eg, SSP-3D), while others do not (\eg, AGORA). In the latter case, we use the YOLOv8~\cite{Jocher_Ultralytics_YOLO_2023} detector (trained on COCO). In the case of AGORA, we evaluate in the fine-tuned setting (NLF fine-tuned just on AGORA), and here we also fine-tune the YOLOv8 detector on AGORA-train.

\paragraph{Evaluation metrics}
In our evaluations, we apply the conventionally used metrics for each individual benchmark. MPJPE is the average Euclidean joint error after alignment at the pelvis. (All distance-based errors are given as millimeters in this work.) Following prior works, \eg \cite{shin2023wham}, on 3DPW and EMDB the reference point is the midpoint between the two hip joints.
P-MPJPE is the Euclidean error after Procrustes alignment, \ie least-squares optimal rigid and uniform scaling alignment.
When 14 joints are indicated for evaluation, these are obtained with a Human3.6M-like regressor from~\cite{kolotouros2019learning}, which is commonly used in the literature.

MVE is analogous to MPJPE but for mesh vertices instead of joints, similarly P-MVE is the Procrustes-aligned version of it.

MPJAE and P-MPJAE are the mean per joint angle errors without and with Procrustes alignment, in degrees. It is the geodesic distance on the $\operatorcall{SO}(3)$ rotation manifold, comparing the global body part orientations between ground truth and prediction.

On SSP-3D, PVE-T-SC evaluates the body shape, by computing the average Euclidean error for the predicted and the ground truth vertices in the default T-pose, after the prediction is optimally scale-aligned to the ground truth. The mIoU is the mean intersection over union metric comparing the binary masks obtained through projecting the ground truth mesh and the predicted mesh on the image plane.

On AGORA, NMVE is a normalized version of MVE that takes into account the detection performance as well. It is the product of MVE and the F1 measure (harmonic mean of precision and recall). Generally, MVE can be improved by not making predictions for hard cases, and the F1 penalty is aimed to compensate for this.

On MPI-INF-3DHP and MuPoTS-3D, PCK stands for the percentage of correct keypoints, namely those predicted joints that are within 150 mm of the ground truth.

\section{Training Details}
\paragraph{Definition of the canonical volume}
The canonical space is defined in reference to a specific T-like pose of the default SMPL mesh. All other keypoints (as e.g. shown in the referenced \cref{fig:results} column 1) are represented in this same coordinate system, as points within this canonical human volume. No explicit conversion between formats is necessary.

\paragraph{Canonical positions for 3D skeletal joints}
The skeleton-based datasets do not provide their correspondence to our SMPL-compatible canonical space.
Hence, we perform an approximate initialization and then treat the canonical position of each of several hundred joints (across the skeleton types) as trainable variables during the main training.

One could obtain this initialization in various ways.
For this work, we first trained a model for predicting the separate skeleton formats (similar to the separate-heads baseline in \cite{Sarandi23WACV}).
We then ran inference with this predictor on the SURREAL dataset and trained linear regressors to interpolate from SURREAL GT vertices to the predicted keypoints and applied this regressor to the canonical template to obtain the approximate initialization.

The corresponding left and right joints are forced to be symmetrically placed in canonical space.
To avoid divergence, we impose a constraint that the canonical positions cannot move away more than 70~mm from their initial positions.

\paragraph{Continuous deformation of body models}
In SMPL(-X), skinning weights are only provided for the discrete set of mesh vertices, so by default forward kinematics can only be applied to these points.
To be able to transform any continuous point (\eg, inside the body), we perform natural neighbor interpolation~\cite{sibson1981brief}.
To avoid having to compute this interpolation on the fly for hundreds of points per training example, we precompute the interpolation weights for $\sim$1 million quasi-randomly sampled points (according to the Sobol sequence).
Then during training, we apply forward kinematics to all SMPL vertices and produce the transformed internal point locations by computing a weighted average of the mesh vertices (according to the pre-computed natural-neighbor interpolation weights).

\paragraph{Point sampling}
Using the above-mentioned interpolation method, we can choose to supervise any set of points in the canonical volume for training examples annotated with high-quality parametric models. We use a total of 1024 points for each such training example, of which 384 are inside the body (sampled from a large, precomputed set of more than a million points) and 640 are on the surface. The surface points are sampled uniformly by picking a random mesh triangle (with probability proportional to the triangle areas) and sampling a random point within the triangle (through barycentric coordinates).
 
\paragraph{Batch normalization}
We found that in our mixture-dataset setting, the batch normalization layers in EfficientNetV2 can accumulate statistics during training that do not work well at inference time.
This applies especially to longer trainings of the large backbone.
We found it effective to use batch renormalization~\cite{Ioffe17NIPS} instead, with default settings. Batch renorm mitigates the discrepancy between the training and inference behavior of batch norm. For numerical stability (\ie, avoiding crashes due to NaN values), we also found that constraining the convolutional kernel norms (to a maximum value of 20) helped.

\paragraph{Training schedule}
We initialize with the 3D pose estimator-based weights from \cite{Sarandi23WACV}, and initially, for 3000 steps, we freeze the image backbone weights and only train the neural field. Subsequently, we decay the backbone learning rate from $3\times 10^{-5}$ to $2\times 10^{-6}$ with exponential decay with a larger drop after 90\% of the training is done.

\paragraph{Augmentations}
We use random rotation, scaling, translation, truncation, color distortion, synthetic occlusion, random erasing and JPEG compression for data augmentation during training.

\section{Inference Speed Optimization}
NLF’s inference-time overhead (for predicting weights through the field MLP) can be eliminated by precomputing the weights once for a chosen set of canonical points. (Typically one wants to localize the same points, i.e. same skeleton formats, for many images.)
For reference, in case of NLF-S with no image batching, and about 8000 points to be predicted (mesh vertices and skeletons), forwarding the field MLP to obtain convolution weights takes 7.7 ms, while the rest of the network including the backbone takes 12.7 ms. For NLF-L with batch size 64 the latter takes 587 ms, making the MLP cost negligible in comparison even if we do not precompute it. 

\section{Uncertainty Estimation}
\begin{table}[tp]
\centering%
\setlength{\tabcolsep}{5pt}%
\titledcaption{Ablation of uncertainty estimation}{%
We measure the average error of joints and vertices of SMPL on the 3DPW benchmark, along with Pearson's correlation (PCC) between the predicted pointwise uncertainty and the true error.
Naively optimizing the log-likelihood harms the mean predictions, but the $\beta_\text{NLL}$ loss~\cite{Seitzer22ICLR} reduces the degradation and improves uncertainty quality. (NLF-S nonparam., on full 3DPW, 24 joints)}
\begin{tabularx}{\linewidth}{@{}lcccc@{}}
\toprule
 & MPJPE$\downarrow$ & MVE$\downarrow$ & PCC-Joints$\uparrow$ & PCC-Vertices$\uparrow$ \\
\midrule
Euclidean loss (no uncertainty) & \B59.2 & \B71.3 & -- & -- \\
Negative log-likelihood (NLL) loss & 61.2 & 73.6 & 0.50 & 0.34 \\
$\beta\text{-NLL}$ loss~\cite{Seitzer22ICLR} ($\beta=1$) & 59.4 & 71.5 & \B0.54 & \B0.40 \\
\bottomrule
\end{tabularx}
\label{tab:uncert}
\end{table}

To also consider our uncertainty prediction $\sigma$ in the loss computation, we can use the negative log-likelihood loss as
\begin{equation}
    \loss{p, NLL}=\frac{\norm{\phat\pr - \p\pr}}{\sigma} + \log \sigma.
\end{equation}

However, we noticed that this significantly diminishes the prediction quality for localization compared to the variant without predicting uncertainties.
We attribute this to the effect described by Seitzer~\etal~\cite{Seitzer22ICLR}, whereby the NLL loss downweights  challenging training examples and impedes the learning progress.
We therefore also experiment with applying their proposed $\beta\text{-NLL}$ loss, which looks as follows when adapted to our context:
\begin{equation}
    \loss{p, \beta-NLL}=\parens{\frac{\norm{\phat\pr - \p\pr}}{\sigma} + \log \sigma}\cdot \operatorcall{sg}{\sigma}^\beta,
\end{equation}
where $\operatorcall{sg}{\cdot}$ is the \enquote{stop gradient} operation that blocks the gradient flow during backpropagation.
Intuitively, for $\beta=1$, multiplying by $\operatorcall{sg}{\sigma}$ \enquote{cancels out} the division of $\norm{\phat\pr - \p\pr}$ by $\sigma$, resulting in the same gradients for $\phat$ as in the case of vanilla Euclidean loss (removing the above-mentioned downweighting of challenging examples), without impacting the gradients for $\sigma$ (due to $\operatorcall{sg}{\cdot}$).

To evaluate these different options, we measure Pearson's correlation coefficient (PCC) between the predicted pointwise uncertainty and the true error for all joints and vertices of the SMPL body model on the 3DPW benchmark test set, using all 24 joints and 6890 vertices.
As seen in \cref{tab:uncert}, the standard NLL loss significantly harms the mean prediction quality, but $\beta_\text{NLL}$ largely fixes this problem and also yields uncertainties that are better correlated with the true error.

\begin{figure}[tp]
\centering
\includegraphics[width=0.5\linewidth]{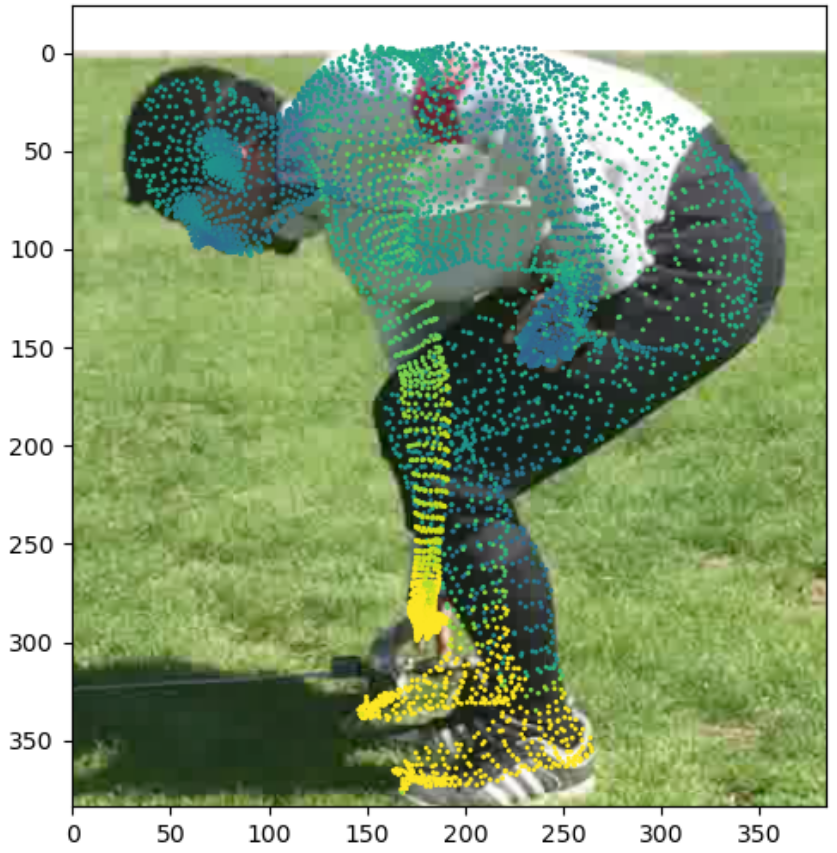} \\
\titledcaption{Uncertainty estimation results}{High uncertainty is indicated in yellow, while low is shown in blue. Occluded body parts tend to have higher uncertainty prediction.}
\label{fig:uncert}
\end{figure}

To illustrate that the obtained uncertainties are visually plausible, we show an example in \cref{fig:uncert}, where the occluded hand has high predicted uncertainty, in line with intuition.

\section{Baseline Architecture Comparison without Localizer Field}
\begin{figure}[tp]
\centering
\begin{tabular}{@{}c@{}}
\includegraphics[width=\textwidth]{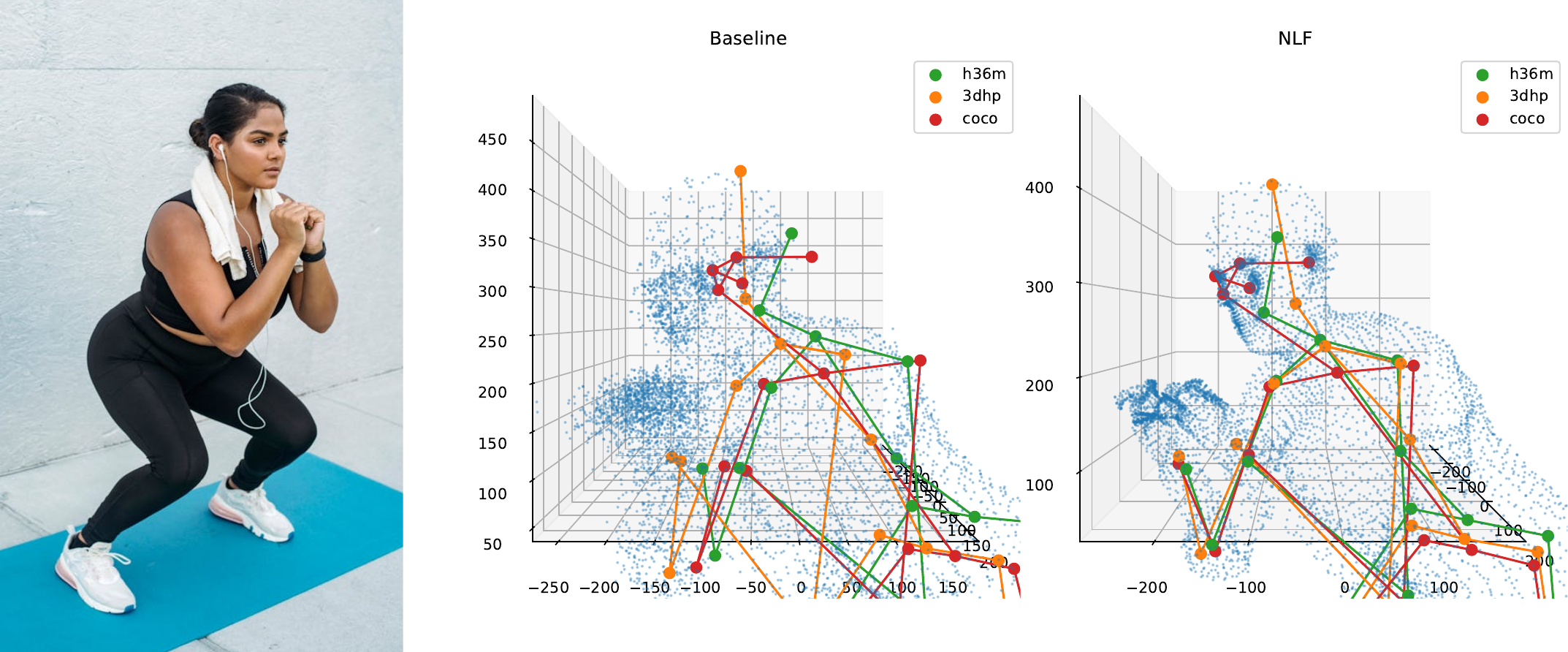}\\
\includegraphics[width=\textwidth]{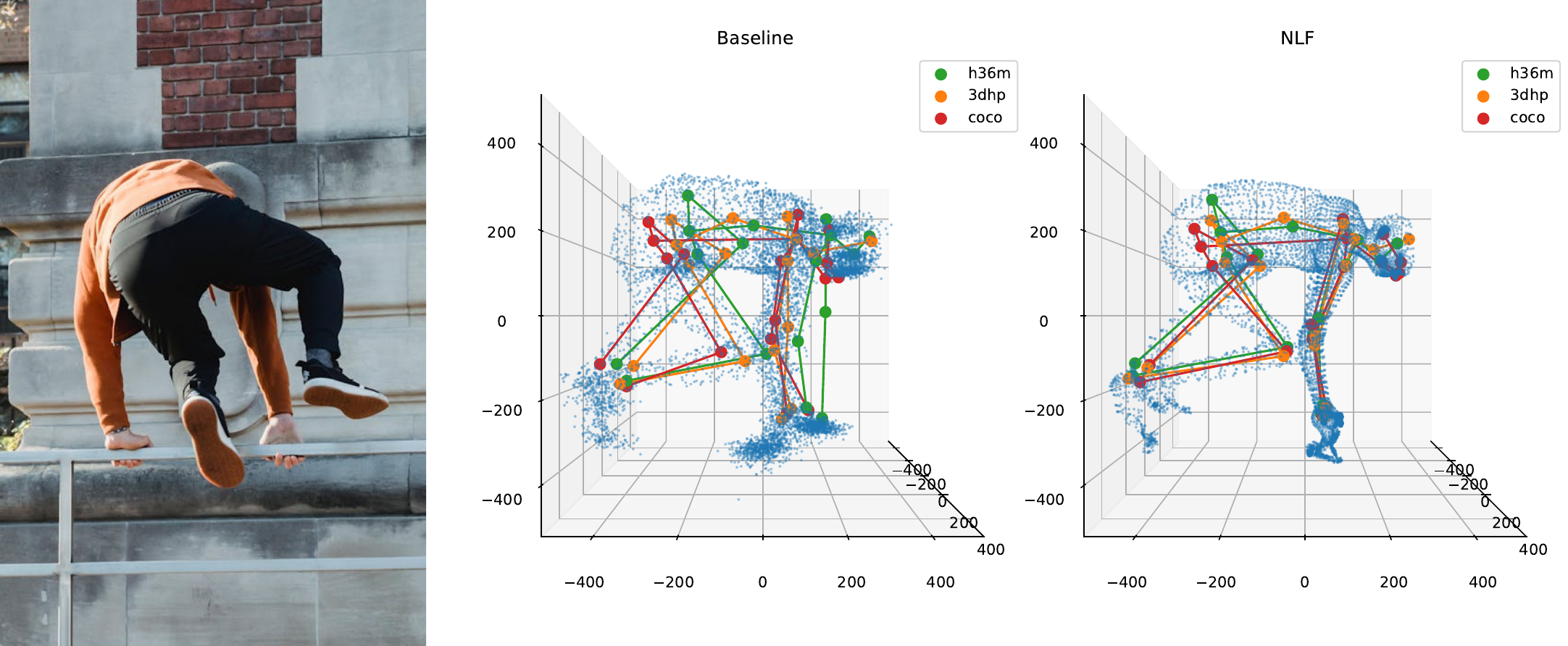}
\end{tabular}
\caption{Side-views of the predictions by our NLF vs. a baseline architecture that trains separate, explicit convolutional weights for each skeletal joint and SMPL vertex.}
\label{fig:separate_weights_baseline}
\end{figure}

To emphasize the key importance of using a localizer field in tying together the heatmap-producing convolutional weights for nearby points of the body, we perform the following ablation.
We create a baseline architecture which uses no localizer field, and instead defines separate learnable parameters for each SMPL vertex and each joint of each skeleton format.

As seen in \cref{fig:separate_weights_baseline}, when using the baseline model for inference, the different skeleton format predictions are visibly inconsistent with each other and with the SMPL mesh (see e.g. the H36M arm outside the SMPL body in the lower example), since the weights to localize each point have no enforced relation to each other. This also results in more scattered vertex predictions (see e.g. the hand region).
NLF, by contrast, ensures that the different skeletons are localized consistently with each other, and the spatial smoothness of the prediction is improved.

\section{Efficient Body Model Fitting}
\begin{figure}[tp]
\centering
\includegraphics[width=\linewidth]{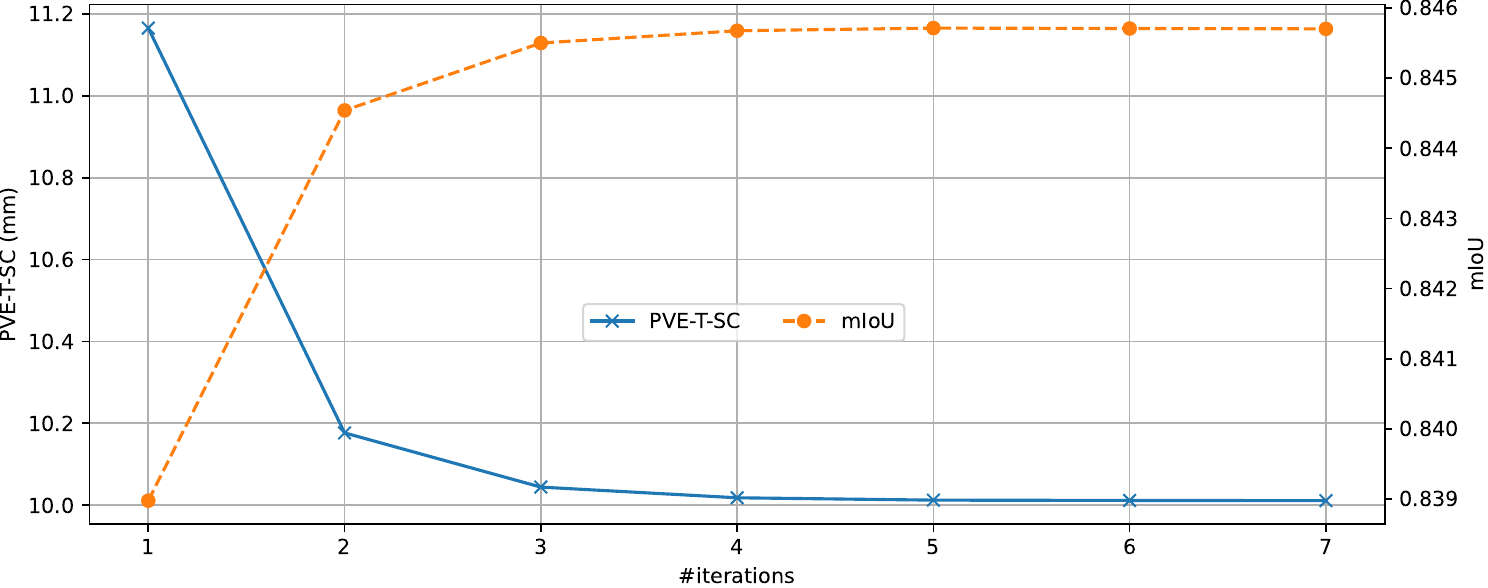}\\
\titledcaption{Convergence of SMPL fitting}{We analyze the convergence properties of our iterative SMPL fitting algorithm on the SSP-3D benchmark.
We use the nonparametric predictions from our NLF-S model as the fitting target. We can observe that approximately three iterations are sufficient for convergence to the SOTA result. (For PVE-T-SC lower is better, and for mIoU higher is better.)}
\label{fig:smpl_fit_convergence}
\end{figure}

\begin{figure}[tp]
\centering
\includegraphics[width=\linewidth]{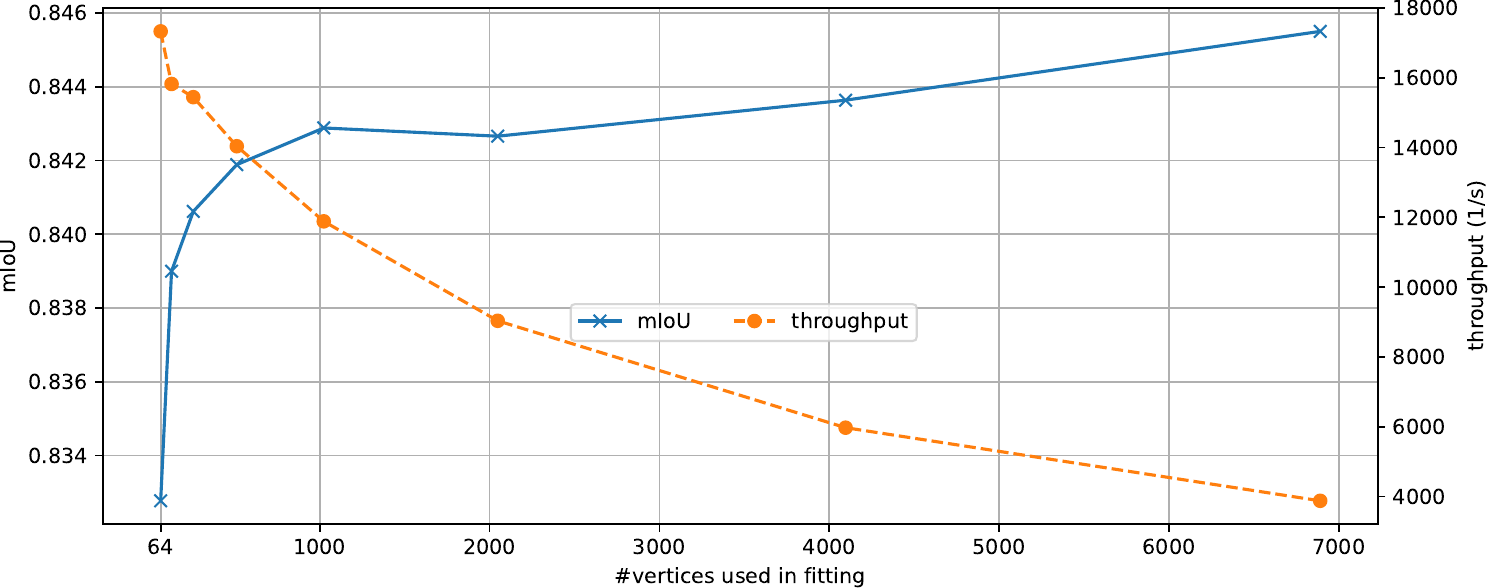}\\
\titledcaption{Vertex-subset-based fitting of SMPL}{For additional efficiency, we evaluate fitting SMPL to only a subset of the 6890 vertices. The joints are still all used. To decide which vertices become part of each subset, we simplify the template mesh by quadric decimation and choose the vertices via distance-based Hungarian matching. We use 3 iterations in all cases.}
\label{fig:smpl_fit_numverts}
\end{figure}

\begin{table}[tp]
\centering
\begin{tabular}{lcccc}
\toprule
Iterations & 25 & 37 & 50 & 100 \\
\midrule
Time & 3 min 18 s & 6 min 54 s & 16 min 35 s & 33 min \\
Error & 14.0 mm & 8.0 mm & 6.2 mm & 5.0 mm \\
\bottomrule
\end{tabular}
\caption{Evaluation of the official SMPL-to-SMPLX converter according to total used time and achieved average error for 33 sample meshes under default settings and various maximum iteration counts. The time required is on the order of \emph{minutes}, while our efficient body model fitter completes the same task with 7.8 mm error in about 28 \emph{milliseconds}.}
\label{tab:fitting_timing}
\end{table}

\begin{table}[tp]
\centering%
\setlength{\tabcolsep}{3pt}%
\titledcaption{Ablation of uncertainty-based weights in body-model fitting}{%
Weighting points according to their estimated uncertainty brings modest benefits on 3DPW and EMDB (NLF-L model).}
\begin{tabularx}{\linewidth}{@{}lcccccccc@{}}
\toprule
 & \multicolumn{4}{c}{3DPW-test (14 joints)} & \multicolumn{4}{c}{EMDB1 (24 joints)} \\
\cmidrule(lr){2-5} \cmidrule(lr){6-9} 
 & MPJPE$\downarrow$  & P-MPJPE$\downarrow$ & MVE$\downarrow$ & P-MVE$\downarrow$ & MPJPE$\downarrow$  & P-MPJPE$\downarrow$ & MVE$\downarrow$ & P-MVE$\downarrow$ \\
\midrule
w/o weighting & 59.3 & 36.6 & \B69.6 & \B48.8 & 68.9 & 41.1 & 81.2 & 51.4 \\
w/ weighting    & \B59.0 & \B36.5 & 69.7 & \B48.8 & \B68.4 & \B40.9 & \B80.6 & \B51.1 \\
\bottomrule
\end{tabularx}
\label{tab:weightedfit}
\end{table}

In \cref{alg:full-fitting}, we provide the simplified pseudocode for our body model fitting algorithm used in the main paper.
The following is an intuitive description of the steps. 
After initialization at the default T-pose, we iterate over the body parts. For each body part, we select the vertices and joints that form part of this body part. (Note that non-endeffector joints belong to two body parts as well.)
We then use the (weighted) Kabsch algorithm to find the rotation matrix between the body part points in the current fit and in the input (fitting target).

Since most joints connect two body parts, it is important that our fit matches the orientation of the bone vectors as closely as possible, so that there is no error accumulation along the kinematic chain.
We therefore weigh the joints much higher than the surface vertices in the rotation estimation.
We specifically set a weight of $10^{-6}$ for vertices.
Even though this weight is tiny, the vertices are meaningfully used as well, since many body parts consist of a single bone with only two joints, and fitting a rotation to two joints would leave one degree of freedom ambiguous (the rotation around the bone's main axis).
By also including the vertices -- albeit with a small weight -- this additional degree of freedom is determined by the vertex positions.
For body parts that contain at least three joints, the rotation fit is based practically solely on the position of these joints (\eg, the pelvis joint is the parent of three joints: two hips and one spine, making this body part contain four joints).
Note that these estimated rotations are not parent-relative.
Each body part's orientation is fitted independently of each other, to avoid any error accumulation.

The body model is then forwarded with the new orientation parameters, yielding new estimated fit vertices and joints, to be used in the second step.

In the second step, the $\beta$ shape parameters and the translation vector $\mathbf{t}$ are estimated.
Since SMPL uses linear blendshapes and linear blend skinning, the vertex and joint positions are a linear function of the shape parameters (while treating the pose rotation parameters as fixed).
We simply need to compute the Jacobian matrix expressing this linear relationship (which involves traversing the kinematic tree).
Given the Jacobian, we can solve for $\beta$ (and $\mathbf{t}$) via L2-regularized linear least squares (solved via Cholesky decomposition).
L2 regularization is used to avoid obtaining unrealistic shape vectors that have too large coefficients.
We found that it is best not to penalize the first two components of $\beta$, which correspond to the person's general size (height and weight), in order to avoid biasing the process to an average-sized shape too strongly.

After obtaining the new shape estimate, we can re-estimate the per-body-part orientations, and this time the body parts have better shape correspondence, hopefully resulting in a more accurate orientation estimation.
With the more accurate orientations, we can re-estimate the shape parameters, and so on.

As a final step, we re-estimate the orientations, but now not independently per body part (anchoring at the mean of each body part), but sequentially, traversing the kinematic chain, anchoring the rotation estimation of each body part at the position of its main joint (as determined by the previous rotation estimations).
Here there is no reason to downweight the contribution of vertices, as the sequential estimation along the kinematic chain ensures that the body parts will link up correctly.
This step reduces misalignment and error accumulation due to mismatched bone lengths between the shape fit and the nonparametrically estimated skeleton.

This process converges very quickly, as shown in \cref{fig:smpl_fit_convergence}, with the main improvement happening between the first and second iterations. From 3 iterations onward, convergence is reached.

To further increase efficiency, we may use only a subset of the vertices of the body model during fitting.
In \cref{fig:smpl_fit_numverts}, we can see that \eg, by fitting to only 1024 vertices, we increase throughput by a factor of 3, while mIoU only decreases from 0.845 to 0.843. The timing was measured on an NVIDIA RTX 3090 GPU with a batch size of 2048.

In the \textbf{model transfer} experiment described in the main paper, where we fit SMPL-X parameters to clean SMPL mesh vertices with our fitting algorithm, we do not regularize the betas (as the input is a clean mesh without noise), use one iteration and 4096 vertices for fitting (out of the 10475). We also provide a more detailed comparison to the official model transfer code in \cref{tab:fitting_timing}, showing that our algorithm is much faster even at the same accuracy levels.

As described in the previous section, our algorithm outputs uncertainty estimates per point.
These can be used to obtain a weighted version of the fitting algorithm.
Specifically, given an estimated uncertainty $\sigma_i$ for point $i$, we use the weighting factors $w_i\propto \sigma_i^{-1.5}$ when solving for the rotations and the shape coefficients.
As shown in \cref{tab:weightedfit}, this gives a slight improvement on both 3DPW and EMDB.

We make all code publicly available (main page at \url{https://istvansarandi.com/nlf}, repository at \url{https://github.com/isarandi/nlf}), including the fitting algorithm (\url{https://github.com/isarandi/smplfitter}), to ensure its precise reproducibility.

\begin{algorithm}[tp]
\newcommand{\templatemesh}{V_0}
\newcommand{\subT}[2]{#1_{#2}^{\mathclap{\hphantom{#2}T}}}
\caption{Efficient fitting of body model $M$ to given target vertices $V$ and joints $J$. $\Pi_{\operatorcall{SO}{3}}$ denotes projection to $\operatorcall{SO}{3}$ via SVD, the subscript $\braces{k}$ selects joints or vertices that belong to body part $k$. $\mathcal{J}$ is the joint regressor. We use $\alpha=10^{-6}$ in practice.}\label{alg:full-fitting}
\begin{algorithmic}
\Require $V \in \R^{N_v\times 3},\ J\in\R^{N_j\times 3},\ I\in \N,\ \alpha \in \brackets{0,1}$ \Comment{}
\State $\tilde{V} \gets \templatemesh,\enspace \tilde{J} \gets \mathcal{J}\tilde{V}, \enspace \mathbf{t} \gets \mathbf{0}, \enspace R_k \gets I_{3\times 3} $ \Comment{Initialize the fit to mean-shaped T-pose}
\For{$i \in \brackets{1..I}$} \Comment{1--4 iterations}
    \LComment{(1) Keeping $\beta$ and $\mathbf{t}$ fixed, solve for global rotations $\mathbf{R}$ via weighted Kabsch}
    \For{$k \in \brackets{1..K}$} 
        \State $P_{\braces{k}} \gets [V_{\braces{k}},\  J_{\braces{k}}]$ \Comment{Concatenate vertices and joints together to a list of points}
        \State $\tilde{P}_{\braces{k}} \gets [\tilde{V}_{\braces{k}},\ \tilde{J}_{\braces{k}}]$ 
        \State $W \gets [\alpha,\ 1-\alpha]$ \Comment{Weights for the vertices and joints, respectively}
        \State $\Sigma_{k} \gets \operatorcall{WeightedCov}{P_{\braces{k}}, \tilde{P}_{\braces{k}}, W}$ \Comment{Weighted cov. mat. with vertices downweighted}
        \State $R_k \gets \Pi_{\operatorcall{SO}{3}}\parens{\Sigma_{k}} R_k$ \Comment{Project to the L2-nearest rotation matrix via SVD}
    \EndFor
    \State $\tilde{V}, \tilde{J} \gets M(\mathbf{R}, \beta) + \mathbf{t}$ \Comment{Pose the vertices and joints with newly fitted $\mathbf{R}$}
    \LComment{(2) Keeping the rotations $\mathbf{R}$ fixed, solve for shape $\beta$ and translation $\mathbf{t}$}
    \State $\tilde{P} \gets [\tilde{V}, \tilde{J}]$ \Comment{Concatenate vertices and joints together for uniform treatment}
    \State $P \gets [V, J]$
    \State $\text{Compute the Jacobian } \nabla_{\beta,\mathbf{t}} \tilde{P} \text{ via forward-mode autodiff. along the kinematic tree}$
    \State $\beta, \mathbf{t} \gets \parens[\big]{\nabla_{\beta,\mathbf{t}} \tilde{P}} \setminus \parens[\big]{P-\tilde{P}}$
    \Comment{Linear least squares (with regularization)}
    \State $\tilde{V}, \tilde{J} \gets M(\mathbf{R}, \beta) + \mathbf{t}$ \Comment{Update vertices and joints with new $\beta$, $\mathbf{t}$}
\EndFor
\LComment{(3) Keeping $\beta$ and $\mathbf{t}$ fixed, refine the rotations $\mathbf{R}$ by traversing the kinematic tree}
\State $\tilde{J}_1^{\text{new}} \gets \tilde{J}_1$ \Comment{Root position is not changed}
\For{$k \in \brackets{1..K}$} \Comment{In topological order of the kinematic tree}
    \State $p \gets \operatorcall{parent-index}{k}$
    \State $\text{For non-root, determine } \tilde{J}_k^{\text{new}} \text{ from } \tilde{J}_p^{\text{new}} \text{, } R_p \text{ and the T-pose k--p bone implied by } \beta$
    \State $P_{\braces{k}} \gets [V_{\braces{k}} - \tilde{J}_k^{\text{new}},\  J_{\braces{k}} - \tilde{J}_k^{\text{new}}]$ \Comment{Get the estimated body part relative to new pivot}
    \State $\tilde{P}_{\braces{k}} \gets [\tilde{V}_{\braces{k}} - \tilde{J}_k,\ \tilde{J}_{\braces{k}}-\tilde{J}_k]$ \Comment{Get the fitted body part relative to its (old) pivot}
    \State $\Sigma_{k} \gets P_{\braces{k}}\phantom{}\T\tilde{P}_{\braces{k}}$ \Comment{\enquote{Covariance} anchored at the respective pivots instead of the means}
    \State $R_k \gets \Pi_{\operatorcall{SO}{3}}\parens{\Sigma_{k}} R_k$
\EndFor \\
\Return $\mathbf{R}, \beta, \mathbf{t}$
\end{algorithmic}
\end{algorithm}%

\section{Neural Field Architecture and Details}
\begin{figure}[tp]
\centering
\includegraphics[width=0.5\linewidth, trim={1cm 18cm 14cm 1cm}, clip=true]{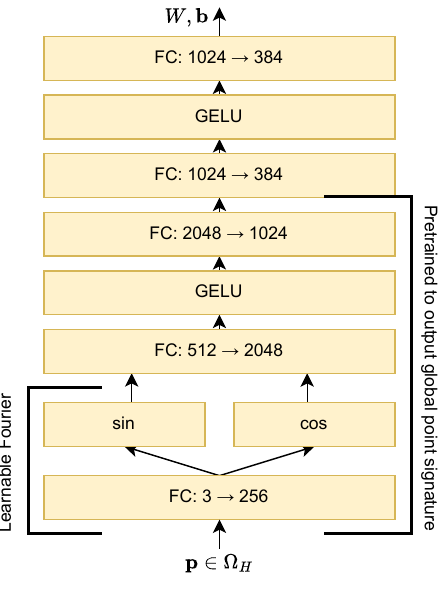}\\
\titledcaption{Architecture of the neural localizer field itself}{}
\label{fig:field}
\end{figure}

\Cref{fig:field} shows the architecture of the neural field that parameterizes our point localizer network. The neural field has an MLP-like structure, starting with learnable Fourier features (fully connected layer followed by sine and cosine activations). After two further fully connected layers with a GELU activation inbetween, we arrive at the layer whose output is initially trained to approximate the global point signature (GPS) derived from the volumetric Laplacian.
Further two FC layers with GELU inbetween yield the parameters to modulate the convolutional layer of the point localizer network as explained in the main paper.

We perform the following steps to obtain the global point signature and bake it into our neural field as initialization.
We take the SMPL mesh in our canonical pose and finely subdivide it with Delaunay tetrahedralization.
We process the resulting tetrahedral mesh with finite element methods to compute the eigenbasis of the Laplacian evaluated at each node of the tet mesh.
These samples become the training set for training an MLP on top of learnable Fourier feature positional encoding.
After this initial setup, the main training commences, where the weights of the neural field are tuned further, together with the backbone net.

\section{Additional 3DPW Protocol}
For fair comparison to~\cite{Sarandi23WACV}, in \cref{tab:3dpw24}, we provide 3DPW evaluations matching their evaluation, \ie, using the entire dataset as the test set (and none of it for training), and evaluating all 24 joints directly (without applying a Human3.6M-like 14-joint regressor).
\begin{table}[tp]
\centering
\setlength{\tabcolsep}{5pt}
\titledcaption{Results on 3DPW (24 joints)}{In contrast to the main paper, here the whole 3DPW dataset used for evaluation, not only the part designated as test set, and all 24 joints are evaluated directly.}
\begin{tabularx}{\linewidth}{@{}lcccc@{}}
\toprule
Method & MPJPE & P-MPJPE & MVE & P-MVE \\
\midrule
MeTRAbs-ACAE-S~\cite{Sarandi23WACV} & 60.6 & 41.7 & -- & --\\
NLF-S & \B59.2 & \B41.0 & 71.7 & 52.0 \\
NLF-S +fit & 59.4 & 41.5 & \B71.1 & \B50.8  \\
\midrule
MeTRAbs-ACAE-L~\cite{Sarandi23WACV} & \B58.9 & \B39.5 & -- & --\\
NLF-L & 59.2 & \B39.5 & 71.0 & 49.2  \\
NLF-L +fit & 59.0 & 40.0 & \B70.1 & \B48.3 \\
\bottomrule
\end{tabularx}
\label{tab:3dpw24}
\end{table}

\section{Further Qualitative Results}
In \cref{fig:qual_supp,fig:qual_supp2} we show further results of NLF-L with SMPL-fitting in post-processing (rendered using the Blendify framework~\cite{blendify2024}).
\newcommand{\addrow}[2]{
  \includegraphics[width=\imagewidth]{figures/qualitative_appendix/pexels-#1_orig.jpg} & 
  \includegraphics[width=\imagewidth]{figures/qualitative_appendix/pexels-#1_overlay.jpg} & 
  \includegraphics[width=\imagewidth]{figures/qualitative_appendix/pexels-#1_rot.jpg} & 
  \includegraphics[width=\imagewidth]{figures/qualitative_appendix/pexels-#2_orig.jpg} & 
  \includegraphics[width=\imagewidth]{figures/qualitative_appendix/pexels-#2_overlay.jpg} &
  \includegraphics[width=\imagewidth]{figures/qualitative_appendix/pexels-#2_rot.jpg} \\
}

\begin{figure}[tp]
\centering
\setlength{\tabcolsep}{1pt} 
\newlength{\imagewidth}
\setlength{\imagewidth}{0.16\linewidth}
\begin{tabular}{cccccc}
\addrow{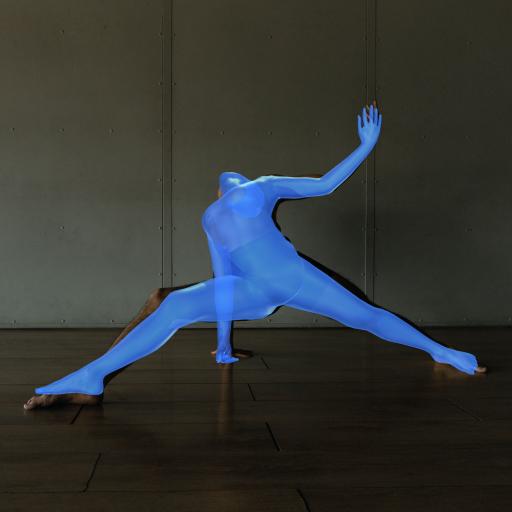}{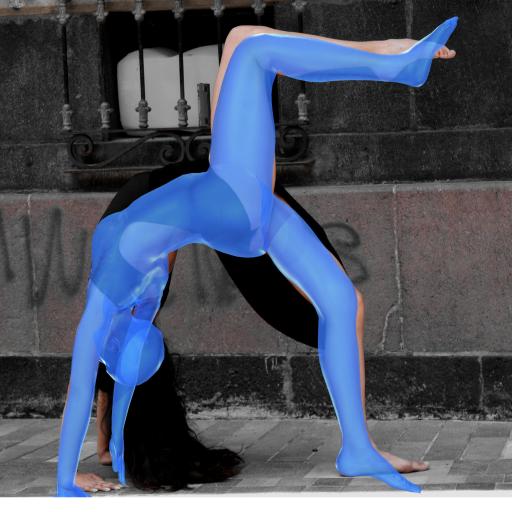}
\addrow{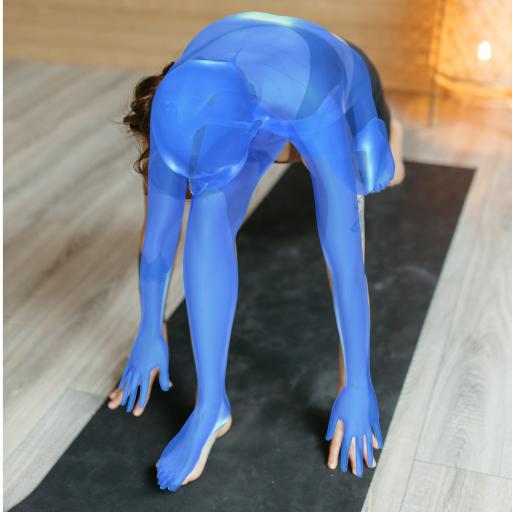}{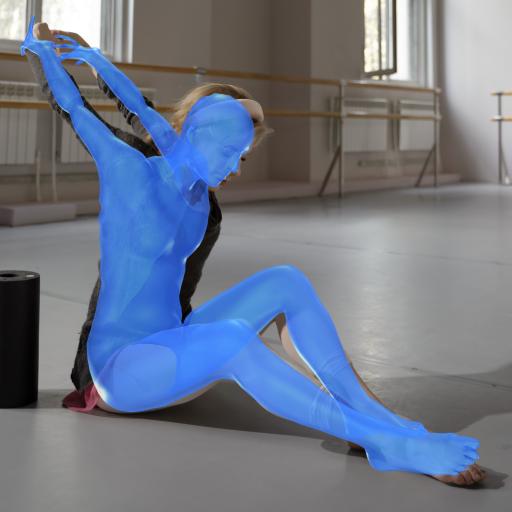}
\addrow{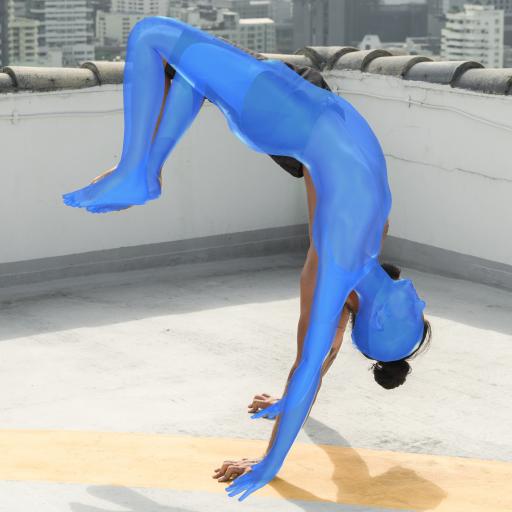}{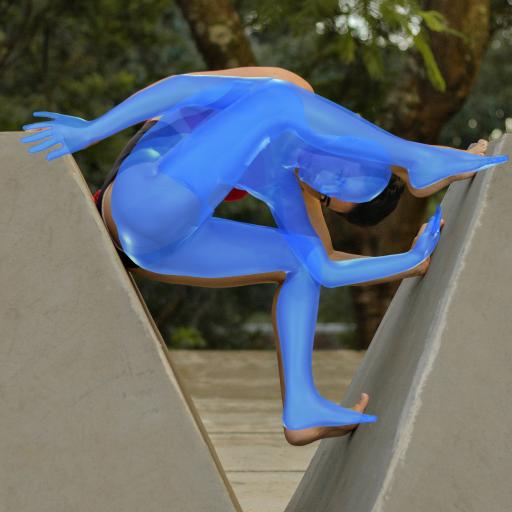}
\addrow{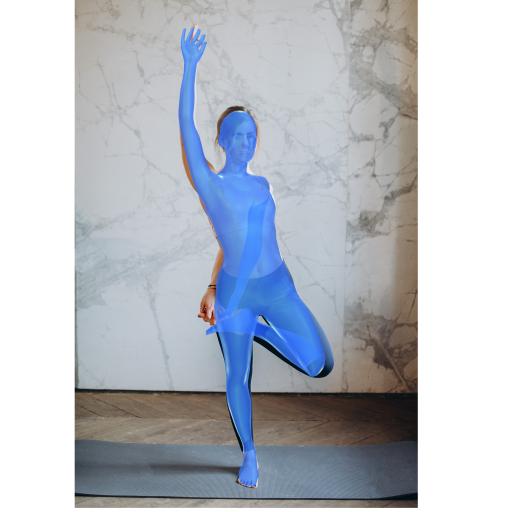}{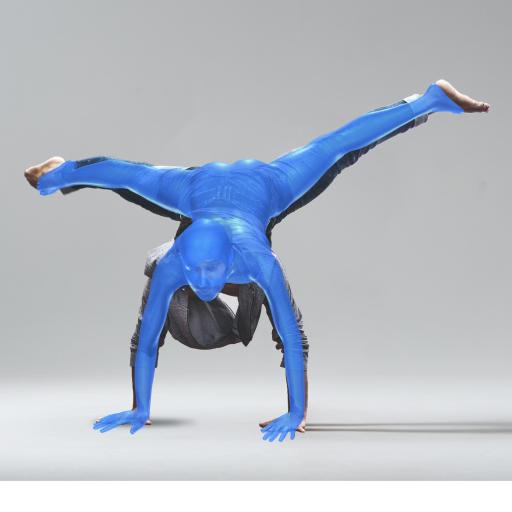}
\addrow{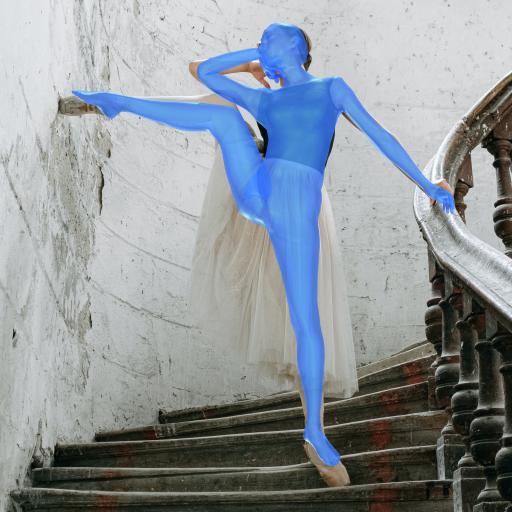}{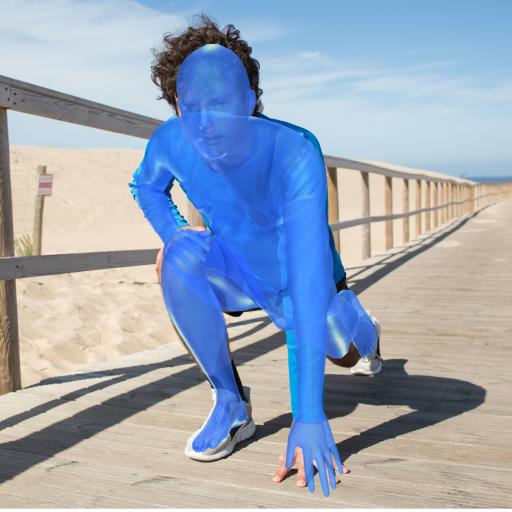}
\addrow{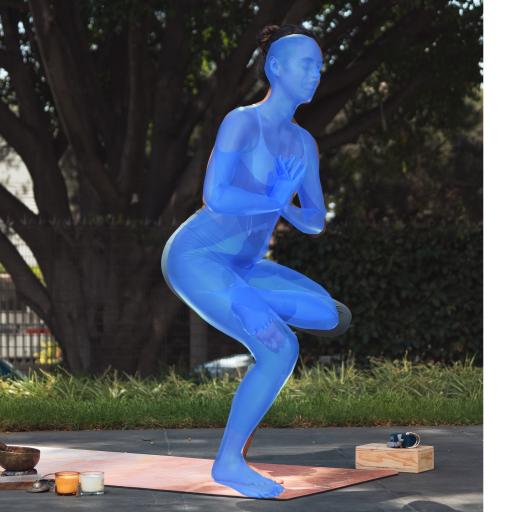}{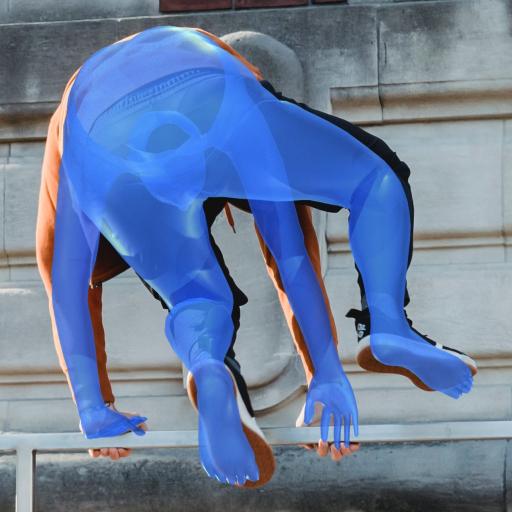}
\end{tabular}
\caption{Qualitative results in the wild.}
\label{fig:qual_supp}
\end{figure}

\begin{figure}[tp]
\centering
\setlength{\tabcolsep}{1pt} 
\setlength{\imagewidth}{0.16\linewidth}
\begin{tabular}{cccccc}
\addrow{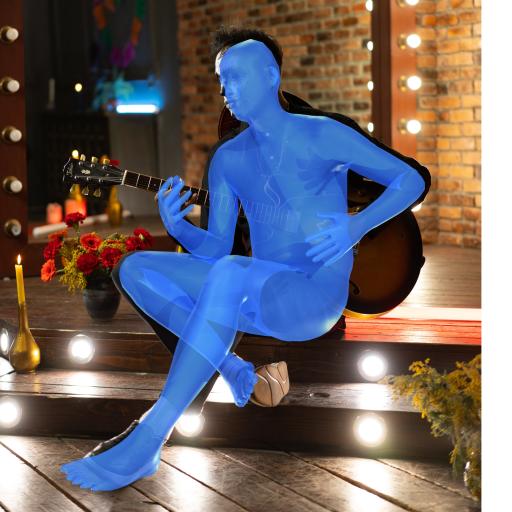}{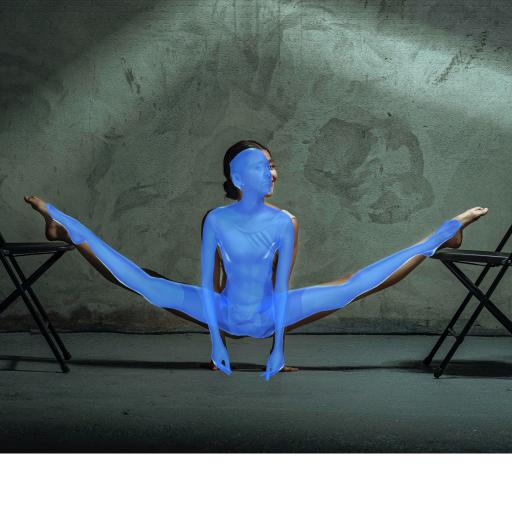}
\addrow{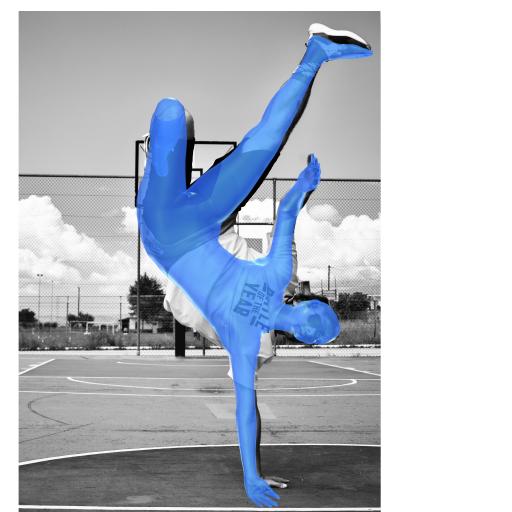}{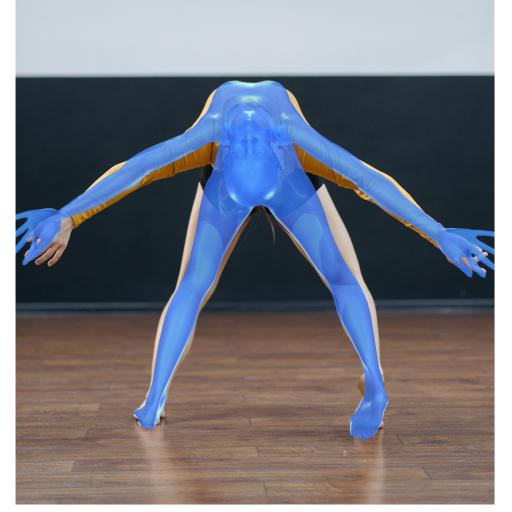}
\addrow{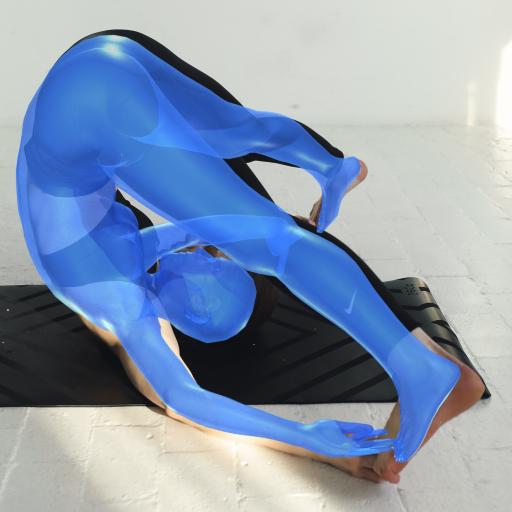}{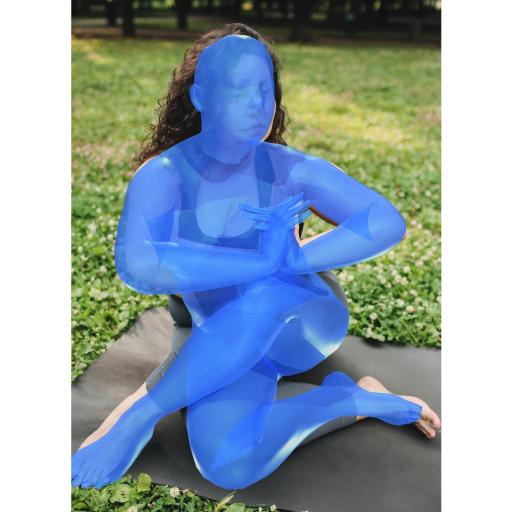}
\addrow{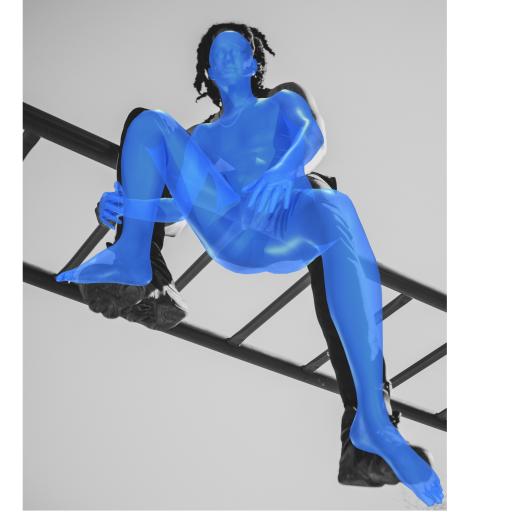}{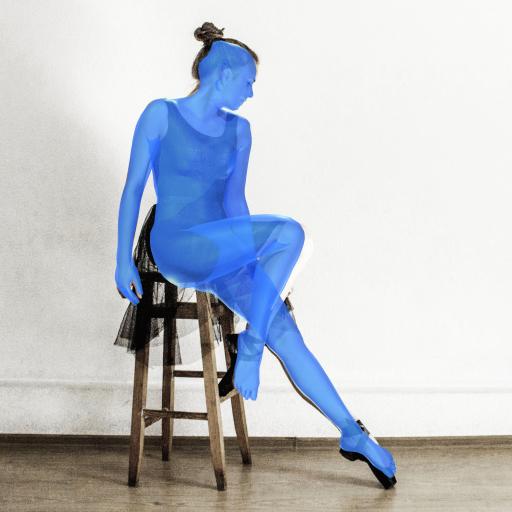}
\addrow{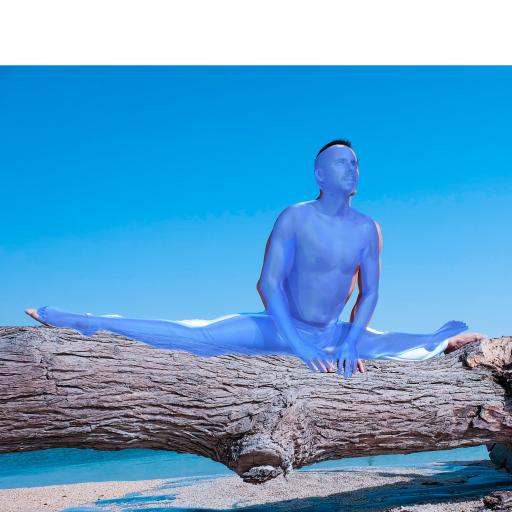}{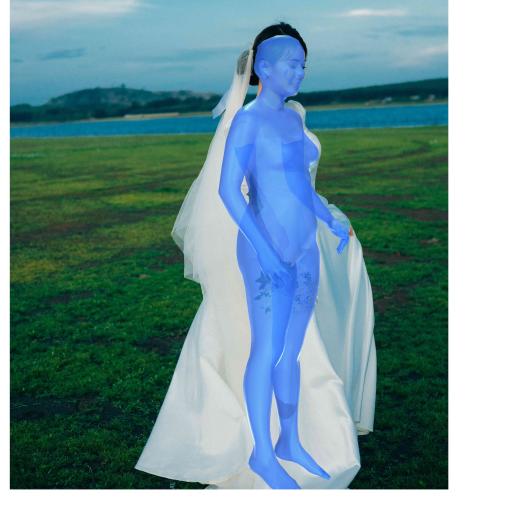}
\addrow{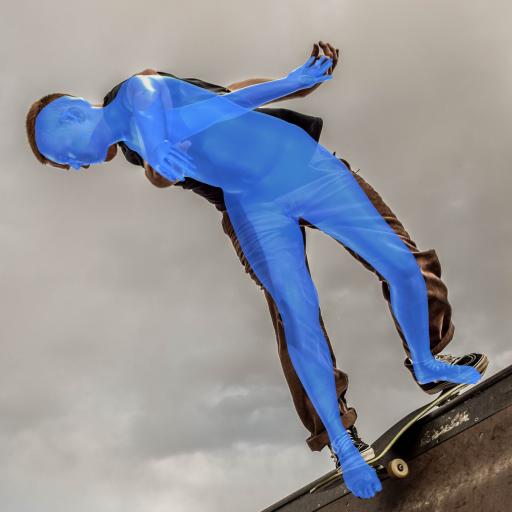}{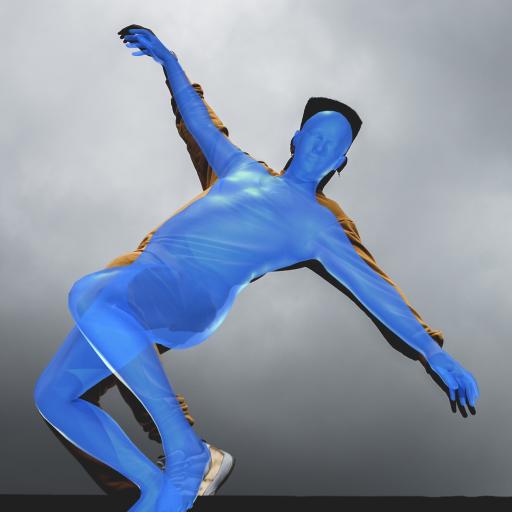}
\end{tabular}
\caption{Qualitative results in the wild.}
\label{fig:qual_supp2}
\end{figure}

\end{document}